\documentclass[runningheads]{llncs} %docclass from the conference

\usepackage[margin=42mm]{geometry}                                  % margins
\usepackage[utf8]{inputenc}                                         % allow utf-8 input
\usepackage[T1]{fontenc}                                            % use 8-bit T1 fonts
\usepackage{graphicx}                                               % images
\usepackage{subcaption}                                             %subfigures
\usepackage{url}                                                    % URL typesetting
\usepackage{booktabs}                                               % good-looking tables
\usepackage{multirow}                                               % for tables
\usepackage{amsfonts}                                               % blackboard math symbols
\usepackage{amsmath}                                                % math ops
\usepackage{nicefrac}                                               % compact 1/2, etc.
\usepackage{microtype}                                              % microtypography
\usepackage{tabularx}
\usepackage{hyperref}

\usepackage{color}

\definecolor{darkblue}{rgb}{0, 0, 0.5}                              % define link colour
\hypersetup{colorlinks=true,citecolor=darkblue,                     % set link colour
            linkcolor=darkblue, urlcolor=darkblue}

\usepackage[natbibapa]{apacite}                                     % only parentheses
\begin{document}
\title{Predicting Depression and Anxiety Risk in Dutch Neighborhoods from Street-View Images}
% old title: {Explainable Deep Learning for the Exploration of Visible Environmental Risk Factors of Depression and Anxiety in The Netherlands}
%
\titlerunning{XDL for Depression\&Anxiety Risk Exploration}
% If the paper title is too long for the running head, you can set
% an abbreviated paper title here
%
\author{Nin Khodorivsko\inst{1}\orcidID{0009-0008-5147-1845} \and
Giacomo Spigler\inst{1}\orcidID{0000-0002-8274-2117} }
\authorrunning{N. Khodorivsko, G. Spigler}
% First names are abbreviated in the running head.
% If there are more than two authors, 'et al.' is used.
%
\institute{1. Tilburg University, Warandelaan 2, 5037 AB Tilburg, The Netherlands}
\maketitle              % typeset the header of the contribution

\begin{abstract}
Depression and anxiety disorders are prevalent mental health challenges affecting a substantial segment of the global population. In this study, we explored the environmental correlates of these disorders by analyzing street-view images (SVI) of neighborhoods in the Netherlands. Our dataset comprises 9,879 Dutch SVIs sourced from Google Street View, paired with statistical depression and anxiety risk metrics from the Dutch Health Monitor.

To tackle this challenge, we refined two existing neural network architectures, DeiT Base and ResNet50. Our goal was to predict neighborhood risk levels, categorized into four tiers from low to high risk, using the raw images. The results showed that DeiT Base and ResNet50 achieved accuracies of 43.43\% and 43.63\%, respectively. Notably, a significant portion of the errors were between adjacent risk categories, resulting in adjusted accuracies of 83.55\% and 80.38\%.

We also implemented the SHapley Additive exPlanations (SHAP) method on both models and employed gradient rollout on DeiT. Interestingly, while SHAP underscored specific landscape attributes, the correlation between these features and distinct depression risk categories remained unclear. The gradient rollout findings were similarly non-definitive. However, through manual analysis, we identified certain landscape types that were consistently linked with specific risk categories. These findings suggest the potential of these techniques in monitoring the correlation between various landscapes and environmental risk factors for mental health issues. As a future direction, we recommend employing these methods to observe how risk scores from the Dutch Health Monitor shift across neighborhoods over time.


\end{abstract}

\section{Introduction} \label{sec:introduction}

Anxiety disorders and depression remain dominant mental health concerns globally \citep{dattani2021}. Traditional methodologies that investigate the relationship between these disorders and environmental aspects like street environments, primarily rely on manual surveys and identification of pre-specified features. These methods are not only expensive but also subjective \citep{odgers2012}. A growing body of recent research underscores the potential of Deep Learning (DL) applied to Street View Images (SVI) as a promising alternative \citep{helbich2019, wang2019, rzotkiewicz2018, biljecki2021}. However, there exists a noticeable gap in exploring this in a Dutch setting, which our study seeks to bridge.

Breaking from the convention of feature extraction dominant in prior studies, we harness the latest advancements in computational power to apply DL directly to SVI. By using Explainable AI (XAI) techniques, our goal is to dissect the patterns the models learn from these images, pinpointing crucial components that play a role in predicting the risk of depression and anxiety.

Our primary research objective is to determine the efficacy of DL techniques applied to SVI in predicting and analyzing the proxy of the incidence of anxiety and depression disorders within the Netherlands. 

To achieve the objective, we fine-tuned two distinct pre-existing neural network architectures. The resulting accuracies achieved non-trivial performance, hinting at the significance of the content of SVIs for predicting mental well-being. Further, analysis using the SHapley Additive exPlanations (SHAP) technique spotlighted specific features as being pivotal in the model's predictive decisions across various categories. Yet, the overlapping influence of these identified features suggests they cannot be earmarked as indicators for any specific class.

Our main contributions are:

\begin{itemize}
\item Collection of a curated Dutch SVI dataset - including both rural and urban areas - with associated labels for "high risk of depression and anxiety" metrics from the Dutch Health Monitor.
\item Fine-tuning two neural network architectures to predict depression and anxiety risk \emph{directly from SVIs}, achieving non-trivial performance.
\item Analysis to identify significant elements in SVIs using Explainable Artificial Intelligence (XAI) techniques.
\item We release the code used to collect the dataset and to perform the analysis at https://github.com/khna89/DL\_streetview\_depression\_anxiety.
\end{itemize}

\section{Related Work}
\subsection{SVI in (mental) health research} 
Street View Imagery (SVI) provides a firsthand look at urban settings, almost as if one is walking on the streets themselves \citep{ki2021, li2015}. Some studies indicate that SVI can provide insights on health-related factors, like how walk-friendly a neighborhood is, in situations where traditional maps or satellite images don't offer clear answers \citep{ki2021}.

Researchers have been tapping into SVI to examine how urban features relate to behavior, health, and demographic patterns \citep{rzotkiewicz2018,biljecki2021}. Recently, deep learning has been employed to identify specific attributes in these images, such as the amount of greenery \citep{ki2021,helbich2019} or types of buildings \citep{nguyen2018}. For example, areas where a lot of sky and trees are visible directly from the street have been linked with lower rates of depression among the elderly \citep{helbich2019,wang2019}. However, these approaches focus on a limited set of features, potentially missing other influential environmental factors. Along these lines, \cite{odgers2012} demonstrated that detecting more features of the environment could be effective in predicting several health-related variables.

In this study, we extend previous work by using machine learning to predict depression and anxiety risk levels \emph{directly from SVIs}, instead of relying on manually defined features. Salient features are then investigated using Explainable AI (XAI) methods.

\subsection{Data collection in SVI-based health studies}
In health studies utilizing Street View Imagery (SVI), the common practice involves measuring the primary metric of interest, such as depression or mobility levels, through individual assessments. Subsequently, researchers extract SVIs around the subjects' residences to aid in model training \citep{helbich2019,ki2021,lu2019}. Fewer studies target aggregate-level data, but a distinguished contribution in this realm is by \cite{keralis2020}. The volume of images acquired per study is varied: while \cite{ki2021} procured an average of 180 SVIs for each participant's location, \cite{helbich2019} amassed approximately 2,807 images per subject, and \cite{lu2019} aggregated around 1,740 images for each housing estate (where multiple participants lived). While the prevalent methodology involves virtually navigating streets to compile images, \cite{keralis2020}, due to the absence of precise residential addresses, opted to determine the coordinates of all junctions within their designated region, leading to the collection of a staggering 31 million images spanning 20,000 neighborhoods. The criteria set for neighborhood selection exhibits differences across studies: \cite{ki2021} selected based on locations provided by survey respondents, \cite{helbich2019} emphasized regions with aging demographics and heightened prevalence of mental disorders, while \cite{lu2019} aimed for a diverse spread in the primary metric but instituted controls for socio-economic variables.

\subsection{Explainability in computer vision}
\label{ssec:expl}

Understanding how a model makes decisions can be crucial, especially when analyzing environmental factors related to mental health. It is thus useful to employ explainability techniques to help make model decisions more transparent and easier to grasp for humans \citep{hase2020, rudin2019}.

One standout method in the realm of Explainable AI (XAI) is the SHapley Additive exPlanations (SHAP) \citep{lundberg2017unified}. While various approaches exist, SHAP brings several of them, like LIME, DeepLIFT, and Layer-wise relevance propagation, under one umbrella. Notably, SHAP is able to align its explanations with human thinking \citep{lundberg2017unified}. Simply put, SHAP values show how much each feature in an image affects the predictions of the  model. In this study, we chose SHAP for its proven human interpretability and model-agnostic nature, enabling its application across our fine-tuned networks.

Additionally, we explored the use of ``attention rollout'', an XAI method specifically designed for vision transformers (ViT) \citep{dosovitskiy2020}. ViTs, while accurate, exhibit complex relationships in their inputs due to their self-attention mechanisms, complicating model interpretation \citep{mahmoudi2023}. The attention rollout technique, as introduced by \citet{abnar2020quantifying}, provides a solution by systematically unrolling attention throughout the model's layers, thereby constructing a comprehensive attention graph. Recent literature indicates the potential of this method in improving the clarity of explanations for image classification tasks \citep{gildenblat2020}.

\section{Methods}

\subsection{Data}

The data used in this project consists of mental health statistics collected per neighbourhood, together with a collection of Street View Images (SVI) sampled for each neighbourhood.

\subsubsection{Mental health data.} Target variable was derived from the Dutch Health Monitor (\cite{gezondheidsmonitor2020}). The Dutch Health Monitor measures depression and anxiety risk based on the Kessler-10 scale \citep{kessler2002}, and contains responses from 540,000 participants aged 18+. Namely, each respondent answered 10 questions, each yielding a score between 1 and 5. The sum of these scores determined the risk level for depression or anxiety disorders. Specifically, a total score of 30-50 categorized a participant as high-risk. The percentage of high-risk individuals in each neighborhood served as our initial continuous target variable. 

We discretized and stratified the risk metrics into 5 levels of risk of depression and anxiety: ``very low'', ``low'', ``moderate'', ``high'', ``very high''. The discretization was accomplished by dividing the range of the ``high risk of depression and anxiety'' scores (in the range of 1\% to 21.9\%) into five uniform intervals (Table \ref{tab:intervals}). This allowed for equally representing depression scores in order to train the classifier to be equally good at all the levels. Nevertheless, this approach does not mirror the actual distribution of the variable, as shown in the Supplementary Materials. Notably, the majority of instances in the monitor fall in the ``very low'' and ``low'' brackets. Since the ``very high'' category was only associated with 7 neighborhood and was thus at risk of overfitting, we decided to combine the ``high'' and ``very high'' categories together into a single class.

\begin{table}
    \caption{Discretization of the variable ``high risk of depression and anxiety'' from the Duth Health Monitor into 5 uniform intervals. The last two intervals were collapsed into one class for further training.}
    \label{tab:intervals}
    \centering
    \small
    \begin{tabular}{|l|l|l|}
    \hline

        Level                & Interval        & Number of neighbourhoods \\
        \hline
        very low             & (1, 5.38)       & 7960 \\
        low                  & (5.39, 9.76)    & 4043 \\
        moderate             & (9.77, 14.14)   & 944 \\
        high \& very high                & (14.15, 18.52) \& (18.53, 22.9)  & 119 (112 \& 7) \\
        \hline
    \end{tabular}
\end{table}

\subsubsection{SVI (General considerations).} Since there is no publicly available Dutch SVI dataset, part of the study involved the collection of the dataset. In order to also cover the gap in studies for rural areas, and taking advantage of the vast farming landscape of the Netherlands, the data was collected from the cities as well as from the rural areas.  

Building upon previous research that highlighted the predictive significance of greenery for mental health variables \citep{wang2019,helbich2019}, our data collection was restricted to the 'green' months, from May to September. This strategy ensures a more accurate representation by mitigating the seasonal variations that might influence the predictive accuracy of each neighborhood on the target variable.

\textbf{Geographically stratified sampling.} To ensure a well-rounded representation of the study area, we devised a sampling protocol to capture a uniform number of images for each category of the target variable. To achieve a balanced dataset, we sampled:

\begin{itemize}
\item 1 image for each ``very low risk'' and ``low risk'' sampled neighborhood;
\item 3 images for each ``moderate risk'' neighborhood;
\item 20 images for each of the combined ``high and very high risk'' neighborhoods.
\end{itemize}

\begin{figure}[h]
    \centering
    \includegraphics[width=0.65\textwidth]{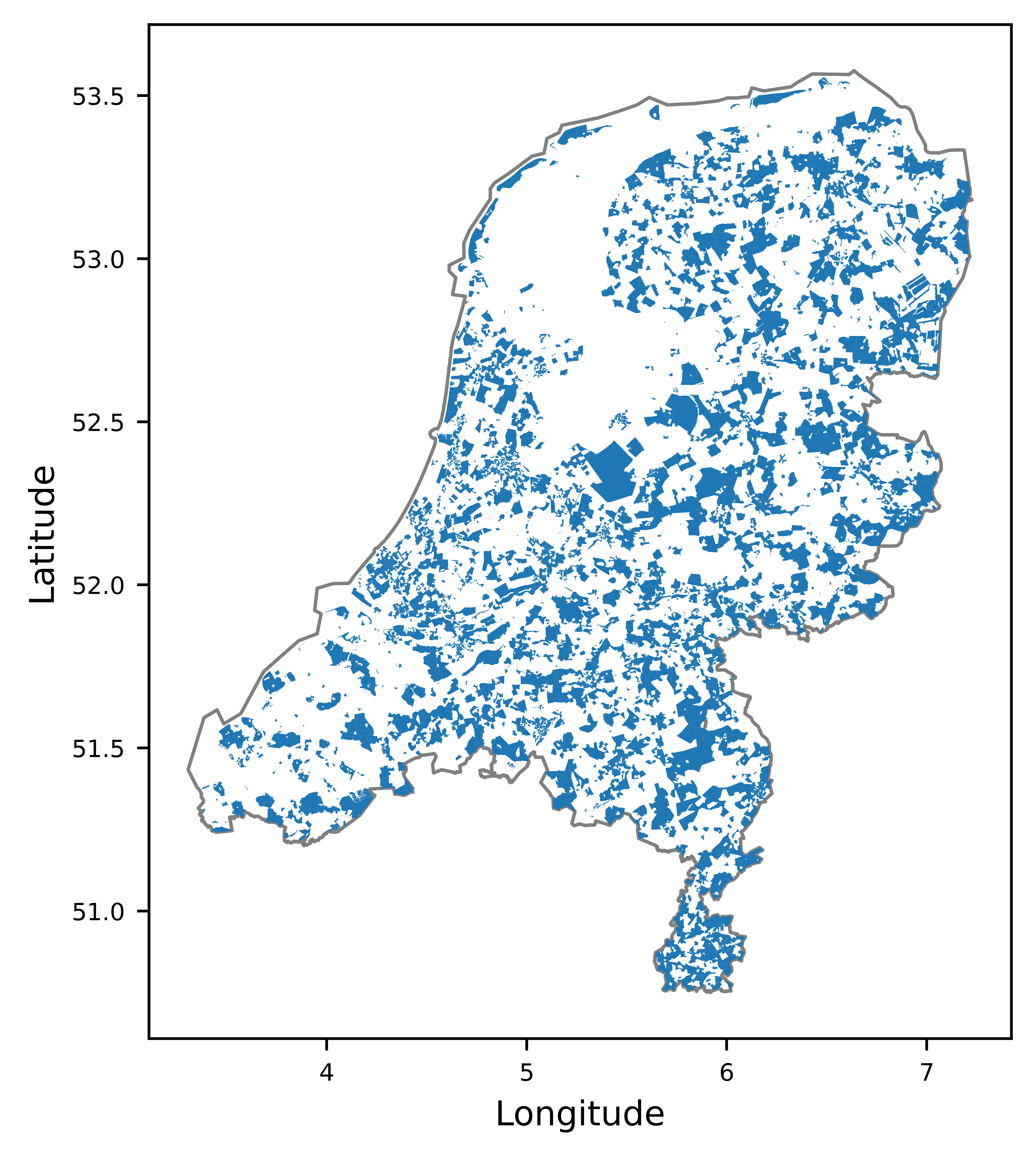} 
    \caption{Geographical distribution of neighbourhoods represented in the dataset.}
    \label{fig:map2}
\end{figure}

More details on this procedure can be found in the Supplementary Materials. From this approach, we amassed a total of 9,879 images (detailed breakdown in Table \ref{tab:table3}), representing 5,952 neighborhoods uniformly distributed across the country (Figure \ref{fig:map2}). This collection method ensures near uniformity over the target classes, albeit with a minor under-representation in the ``high and very high'' category.

\begin{table}[h]
    \centering
    \small
    \caption{Number of street-view images for each target class.}
    \label{tab:table3}
    \centering
    \small
    \begin{tabular}{|l|l|}
        \hline
        Target class & Number of images \\
        \hline
        very low               & 2500     \\
        low                    & 2500     \\
        moderate               & 2499     \\
        high and very high     & 2380     \\
        \hline
    \end{tabular}
\end{table}

\subsubsection{SVI (Data preprocessing).} The street-view images were randomly cropped from 512x512 pixels to 224x224 pixels for training to prevent overfitting and to match the input size expected by pre-trained models. For validation and testing, full images were resized to 224x224 pixels. Image RGB channels were normalized using ImageNet statistics, $\text{mean}=(0.485, 0.456, 0.406), \text{std}=(0.229, 0.224, 0.225)$. The dataset was randomly partitioned into training (70\%), validation (15\%), and test (15\%) sets.

%{\color{blue} \textbf{EXPLAIN THE SPLIT}} {\color{red} \textbf{the long version would be:} A random split was chosen for several reasons. First, for the lowest two risk categories, each neighborhood contributes only one image, making a per-neighborhood split impractical. Second, for the third risk level, each neighborhood contributes three images taken from random locations within the neighborhood, reducing the likelihood of the model memorizing specific neighborhoods. Lastly, for the highest risk category, which has 20 images per neighborhood, the model's performance on the test set suggests that overfitting to specific neighborhoods is unlikely. Specifically, ResNet performs better on this category than on the second and third risk levels, indicating generalization rather than memorization. Additionally, the random split makes it improbable that all 20 images from a single neighborhood would be allocated to just one of the sets. Therefore, we deemed a random split to be a suitable approach for partitioning the data.}.

\subsection{Experimental setup}
\label{ssec:exp_setup}

Two pre-trained models, DeiT \citep{touvron2021} and ResNet-50 \citep{he2016}, were used as base for the experiments. The models were modified by replacing the output layer, specific for the classification task of interest over 4 categories. A cross-entropy loss was used.

Hyperparameter tuning was performed using grid search on a subset of 30\% of the dataset for training and 10\% for validation. For DeiT, we explored different architectures (Tiny, Small, Base), number of unfrozen layers (0, 1, 3, 5), learning rates (0.01, 0.005, 0.001), and optimizers (Adam, SGD, Adagrad). For ResNet-50, similar parameters were tested, with the addition of the RMSprop optimizer. Both models were then fine-tuned using the best hyperparameter combinations and the full training set. Detailed hyperparameter values are provided in the Supplementary Materials.

Fine-tuning was performed by trainng the models for up to 100 epochs with early stopping triggered if validation loss did not improve for 10 epochs. In practice, early stopping was always triggered before reaching 100 epochs. Regularization techniques were used to reduce risk of overfitting, including dropout, L2 regularization, learning rate scheduling, and early stopping.

\subsection{Evaluation and Explainability}
Evaluation and explainability were conducted on held-out test data.

\textbf{Performance evaluation.} Model performance was assessed via accuracy, average loss, and weighted F1 scores. Additionally, we introduced an `adjusted accuracy' metric that ignores misclassifications across adjacent classes, instead considering them correct. Given the continuous character of the target variable and the steps taken in its discretization, this metric holds significant relevance. For a deeper understanding of its importance, refer to the \hyperref[sec:disc-concl]{Discussion} section.

\textbf{Model explanation.} To delve deeper into the model's mechanisms, we selected samples that were correctly classified. The samples were then ranked based on the outputs of the models, and the 10 samples with the highest predicted probabilities from each class were earmarked for manual exploration and further test with XAI methods.

For the computation of SHAP values, we established a background dataset comprising 50 random images using the GradientExplainer. To visualize importance scores, we tweaked the `shap.image\_plot' function to facilitate individual file exports.

Finally, we applied Gildenblat's 2020 version of attention rollout to study the final DeiT model. This tool allows for the visualization of attention weights, offers options for displaying extreme weights selectively, and lets users choose a fusion method. We set a discard ratio at 0.8, enhancing the clarity of our visualizations. For additional details, we refer to the Supplementary Materials.

\section{Results}

The performance of the trained models is reported in Table \ref{tab:table4}. Both the transformer (DeiT) and CNN (ResNet) models achieved moderate performance, with ResNet slightly surpassing DeiT. 

\begin{figure}[h]
    \centering
    \begin{subfigure}{.5\textwidth}
        \centering
        \includegraphics[width=.9\linewidth]{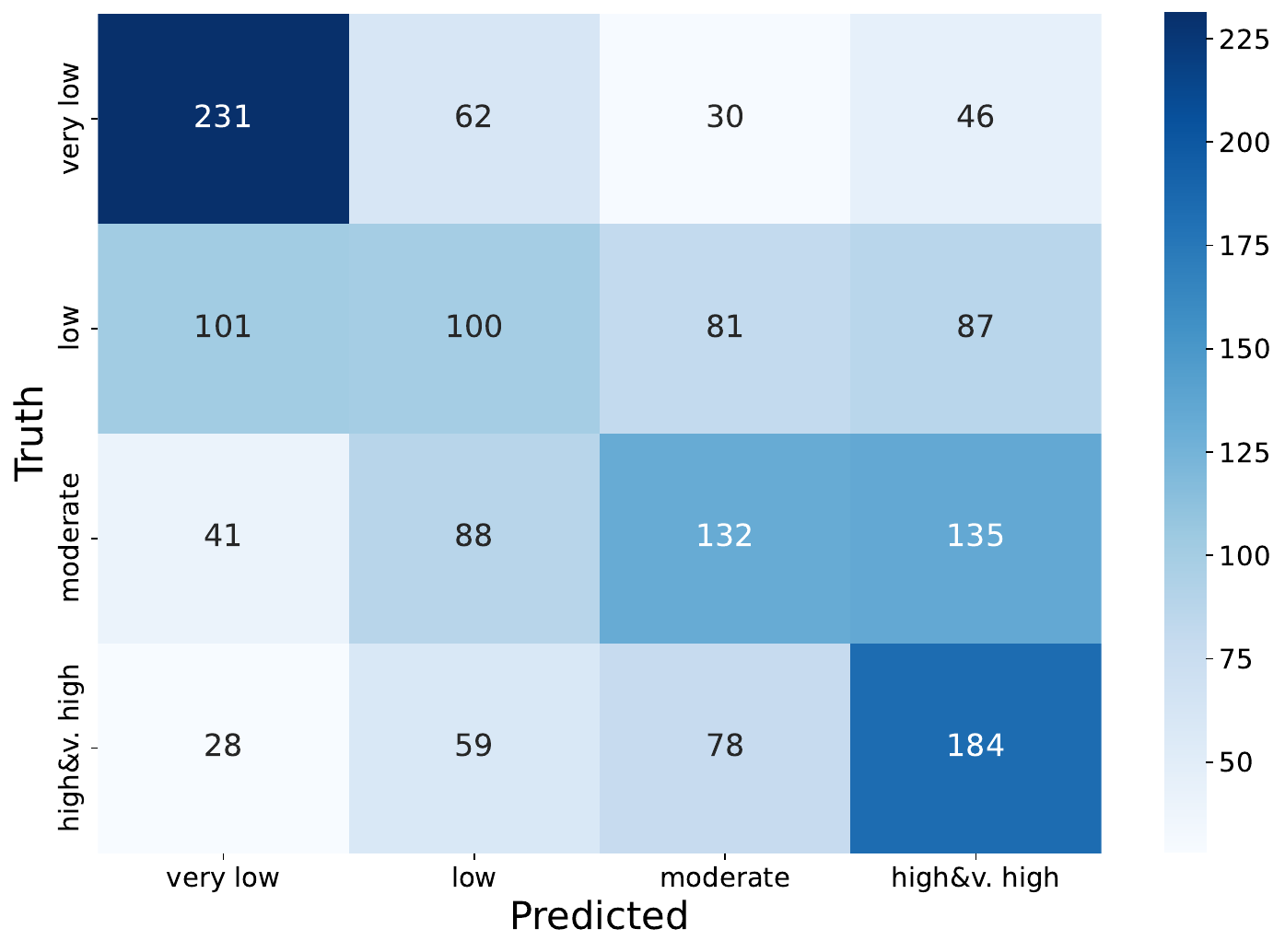}
        \caption{}
        \label{fig:conf1}
    \end{subfigure}%
    \begin{subfigure}{.5\textwidth}
        \centering
        \includegraphics[width=.9\linewidth]{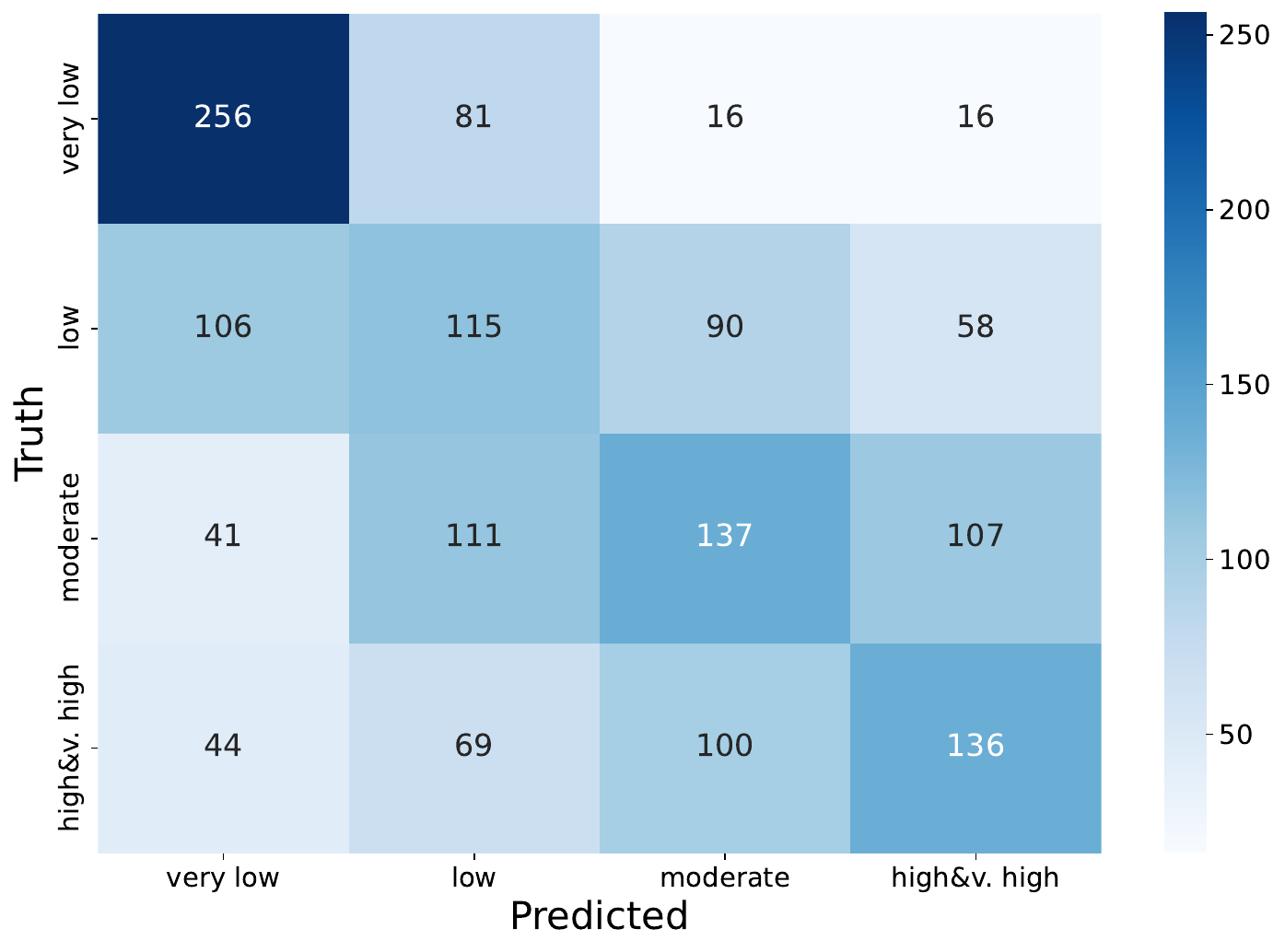}
        \caption{}
        \label{fig:conf2}
    \end{subfigure}
    \caption{Confusion matrices of the predictions of fully fine-tuned models. a) ResNet50. b) Deit Base.}
    \label{fig:conf}
\end{figure}

As shown in the confusion matrices (Fig. \ref{fig:conf}), most misclassifications are found within adjacent classes. This is due to the discretization of the original continuous target variable, for which two similar samples, e.g., 5.38 and 5.39, which lie just across the threshold between the ``very low'' and ``low'' categories, are assigned to two distinct categories. We thus adopt an `adjusted accuracy' metric that accepts as correct all classifications within adjacent classes.

\begin{table}[h]
    \centering
    \small
    \caption{Test-set performance of trained models. Chance-level is $25\%$, with 62.5\% for adjusted accuracy.}
    \label{tab:table4}
    \begin{tabular}{|l|r|r|r|r|}
        \hline
        Model & Test accuracy & Adjusted accuracy & Test F1 & Test loss \\
        \hline
        Deit Base  & 0.43 & 0.84 & 0.44 & 1.21 \\
        ResNet-50  & 0.44 & 0.80 & 0.44 & 1.31 \\
        \hline
    \end{tabular}
\end{table}

\subsection{ResNet50}

The best model selected through hyperparameter tuning had 3 unfrozen layers, Adagrad optimizer, learning rate of 0.01, two dropout layers with drop probability of 0.1, and L2 regularization with $\lambda = 0.00005$. The model achieved an accuracy of 43.63\% on the test set, with an adjusted accuracy of 80.38\% (Table \ref{tab:table4}).

\begin{figure}[htbp]
    \centering

    % Row for class "very low"
    \caption*{\textbf{very low risk:}}

    \begin{subfigure}[b]{0.24\textwidth}
        \includegraphics[width=\textwidth]{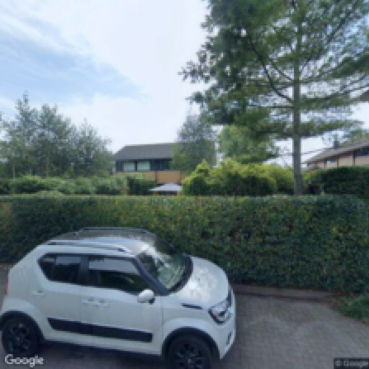}
        \caption*{}
    \end{subfigure}
    \begin{subfigure}[b]{0.24\textwidth}
        \includegraphics[width=\textwidth]{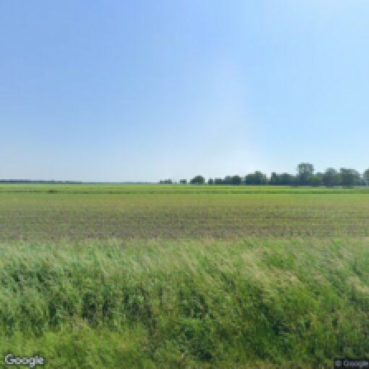}
        \caption*{}
    \end{subfigure}
    \begin{subfigure}[b]{0.24\textwidth}
        \includegraphics[width=\textwidth]{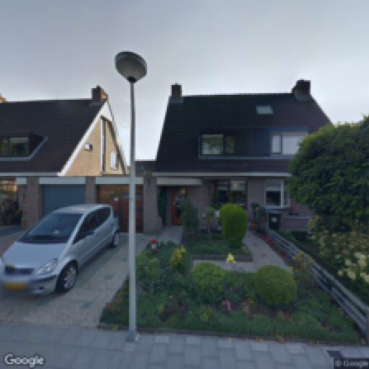}
        \caption*{}
    \end{subfigure}
    \begin{subfigure}[b]{0.24\textwidth}
        \includegraphics[width=\textwidth]{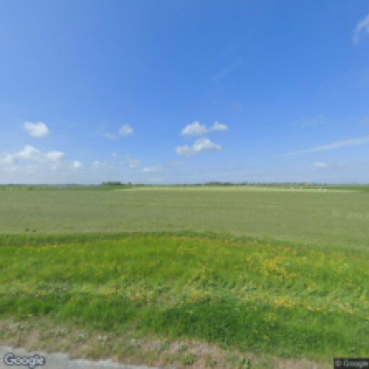}
        \caption*{}
    \end{subfigure}

    % Row for class "low"
    \caption*{\textbf{low risk:}}
    \begin{subfigure}[b]{0.24\textwidth}
        \includegraphics[width=\textwidth]{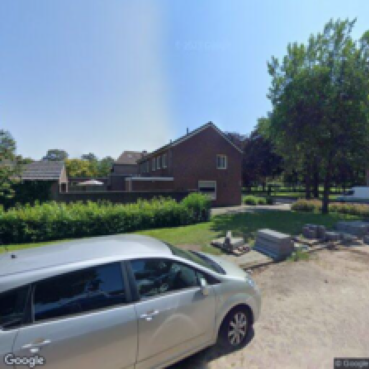}
        \caption*{}
    \end{subfigure}
    \begin{subfigure}[b]{0.24\textwidth}
        \includegraphics[width=\textwidth]{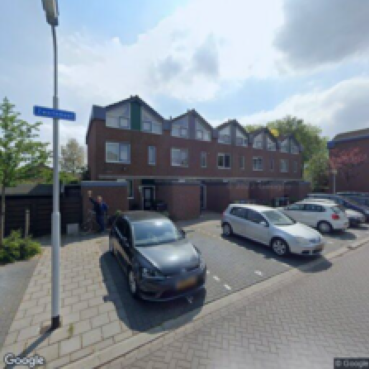}
        \caption*{}
    \end{subfigure}
    \begin{subfigure}[b]{0.24\textwidth}
        \includegraphics[width=\textwidth]{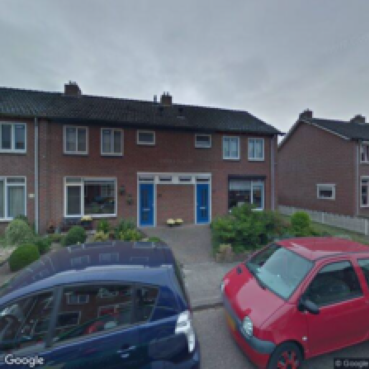}
        \caption*{}
    \end{subfigure}
    \begin{subfigure}[b]{0.24\textwidth}
        \includegraphics[width=\textwidth]{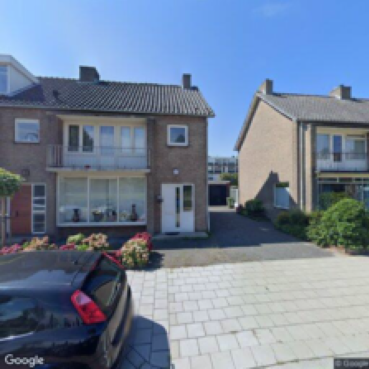}
        \caption*{}
    \end{subfigure}

    % Row for class "moderate"
    \caption*{\textbf{moderate risk:}}

    \begin{subfigure}[b]{0.24\textwidth}
        \includegraphics[width=\textwidth]{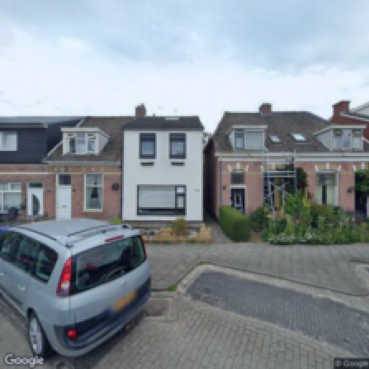}
        \caption*{}
    \end{subfigure}
    \begin{subfigure}[b]{0.24\textwidth}
        \includegraphics[width=\textwidth]{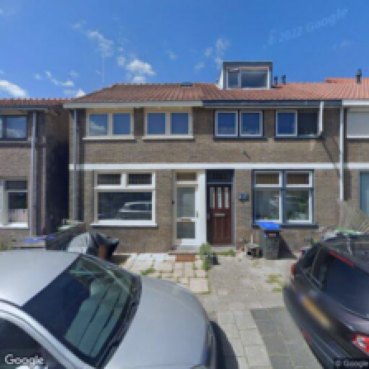}
        \caption*{}
    \end{subfigure}
    \begin{subfigure}[b]{0.24\textwidth}
        \includegraphics[width=\textwidth]{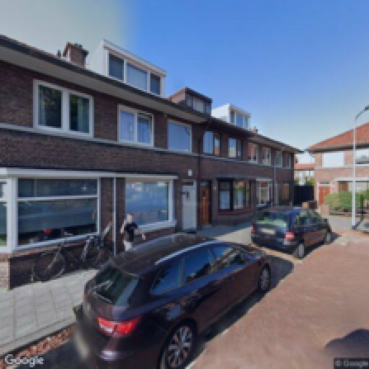}
        \caption*{}
    \end{subfigure}
    \begin{subfigure}[b]{0.24\textwidth}
        \includegraphics[width=\textwidth]{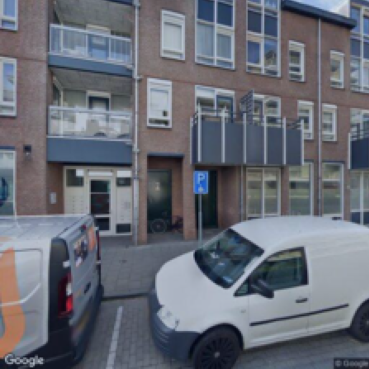}
        \caption*{}
    \end{subfigure}

    \caption*{\textbf{high, very high risk:}}
    \begin{subfigure}[b]{0.24\textwidth}
        \includegraphics[width=\textwidth]{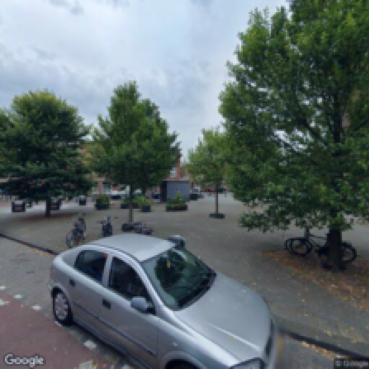}
        \caption*{}
    \end{subfigure}
    \begin{subfigure}[b]{0.24\textwidth}
        \includegraphics[width=\textwidth]{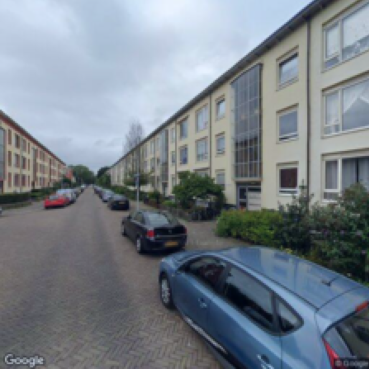}
        \caption*{}
    \end{subfigure}
    \begin{subfigure}[b]{0.24\textwidth}
        \includegraphics[width=\textwidth]{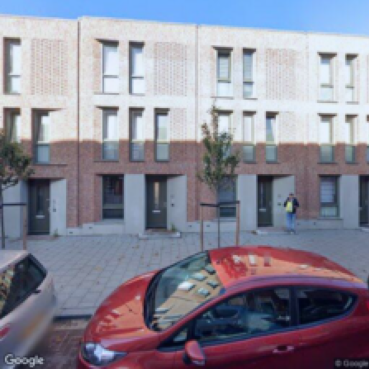}
        \caption*{}
    \end{subfigure}
    \begin{subfigure}[b]{0.24\textwidth}
        \includegraphics[width=\textwidth]{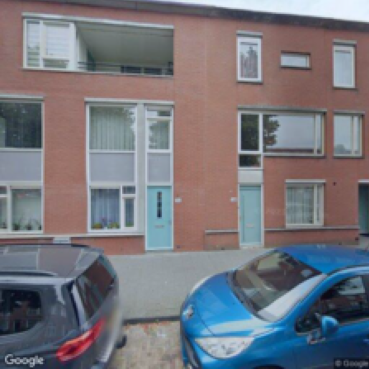}
        \caption*{}
    \end{subfigure}
    
    \caption{SVIs that were correctly identified by the fine-tuned ResNet50 as pointing to the given level of depression and anxiety risk, per level of the target variable. Images: Google Street View.}
    \label{fig:best_resnet}
\end{figure}

Our analysis of the true positive predictions with the highest logits uncovered consistent patterns across different risk levels, as illustrated in Fig. \ref{fig:best_resnet}. Specifically:
\begin{itemize}
\item The ``very low'' risk class mostly displayed rural landscapes, often devoid of built structures but occasionally featuring single-family homes.
\item ``Low risk'' images typically portrayed terraced houses, with the occasional sight of duplexes or single-family homes. These images often had clear skies and cultivated greenery.
\item The ``moderate" as well as the ``high and very high'' levels frequently presented blocks of flats and cars, with limited greenery and sky in view.
\end{itemize}

Our exploratory analysis using the SHapley Additive exPlanations (SHAP) method uncovered features that were ambiguous in nature, such as expansive sky regions and cars, which influenced model predictions across various classes. These features displayed both positive and negative SHAP values, indicating their non-distinctive role in determining the risk levels for depression and anxiety. Notable among these ambiguous features are vast sky areas near the horizon, roofs, shadows beneath roofs, shaded walls, tree canopies, and cars. Efforts to improve clarity by deducting mean SHAP values for incorrect classifications were unsuccessful. Visual representations of these SHAP values can be found in the {Supplementary Materials}.

\subsection{DeiT}

The best DeiT model selected through hyperparameter tuning utilized 5 unfrozen layers, SGD optimizer, a learning rate of 0.001, dropout layers with drop probability of 0.2, and \( L_2 \) regularization with (\( \lambda = 0.0001 \)). The model achieved 43.43\% accuracy on the test set, with adjusted accuracy of 83.55\%, and was thus comparable in performance to the ResNet50 model.

\begin{figure}[htbp]
    \centering

    % Row for class "very low"
    \caption*{\textbf{very low risk:}}
    \begin{subfigure}[b]{0.24\textwidth}
        \includegraphics[width=\textwidth]{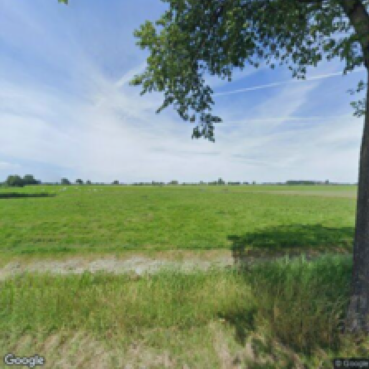}
        \caption*{}
    \end{subfigure}
    \begin{subfigure}[b]{0.24\textwidth}
        \includegraphics[width=\textwidth]{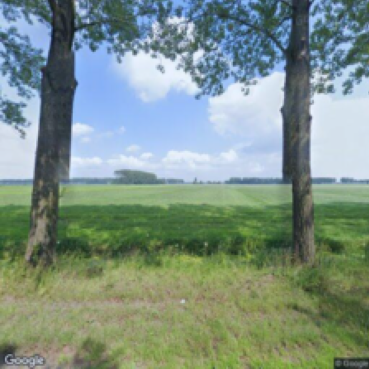}
        \caption*{}
    \end{subfigure}
    \begin{subfigure}[b]{0.24\textwidth}
        \includegraphics[width=\textwidth]{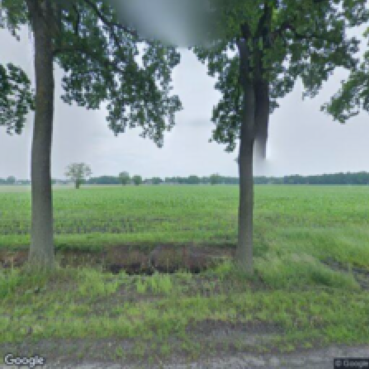}
        \caption*{}
    \end{subfigure}
    \begin{subfigure}[b]{0.24\textwidth}
        \includegraphics[width=\textwidth]{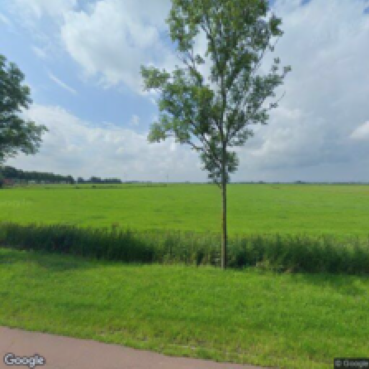}
        \caption*{}
    \end{subfigure}

    % Row for class "low"
    \caption*{\textbf{low risk:}}
    \begin{subfigure}[b]{0.24\textwidth}
        \includegraphics[width=\textwidth]{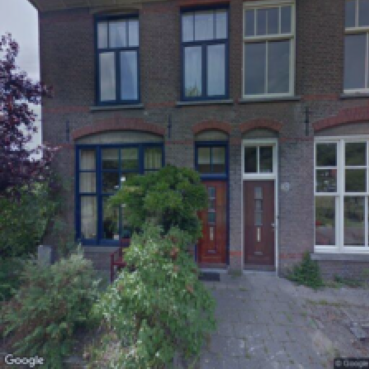}
        \caption*{}
    \end{subfigure}
    \begin{subfigure}[b]{0.24\textwidth}
        \includegraphics[width=\textwidth]{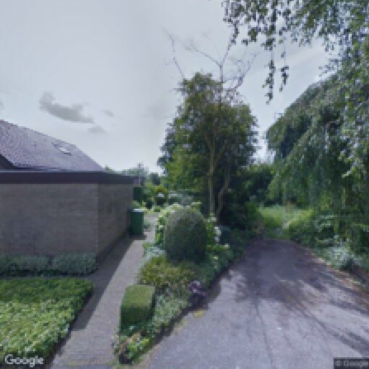}
        \caption*{}
    \end{subfigure}
    \begin{subfigure}[b]{0.24\textwidth}
        \includegraphics[width=\textwidth]{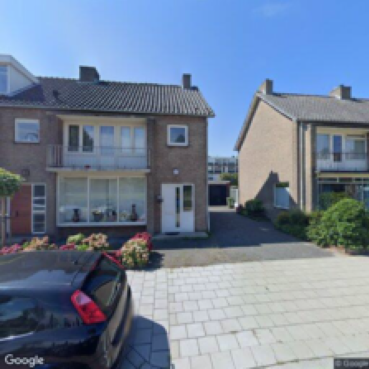}
        \caption*{}
    \end{subfigure}
    \begin{subfigure}[b]{0.24\textwidth}
        \includegraphics[width=\textwidth]{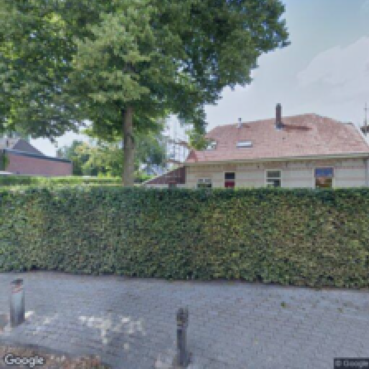}
        \caption*{}
    \end{subfigure}

    % Row for class "moderate"
    \caption*{\textbf{moderate risk:}}
    \begin{subfigure}[b]{0.24\textwidth}
        \includegraphics[width=\textwidth]{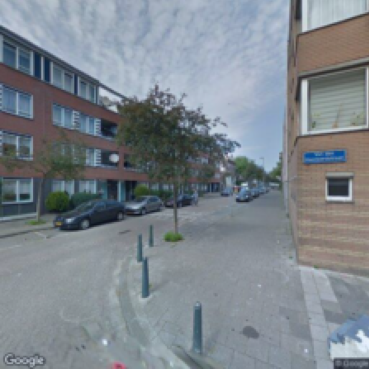}
        \caption*{}
    \end{subfigure}
    \begin{subfigure}[b]{0.24\textwidth}
        \includegraphics[width=\textwidth]{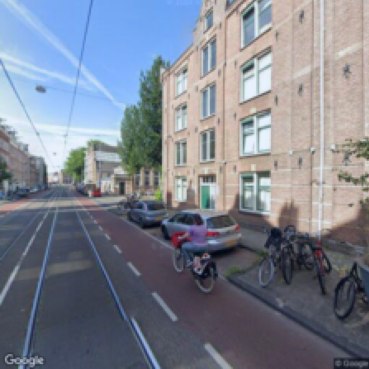}
        \caption*{}
    \end{subfigure}
    \begin{subfigure}[b]{0.24\textwidth}
        \includegraphics[width=\textwidth]{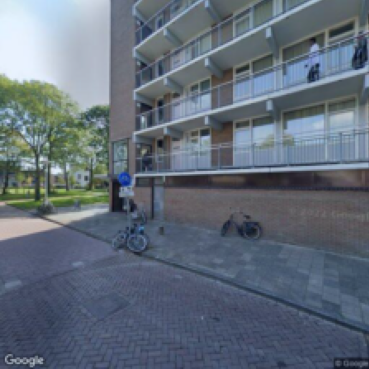}
        \caption*{}
    \end{subfigure}
    \begin{subfigure}[b]{0.24\textwidth}
        \includegraphics[width=\textwidth]{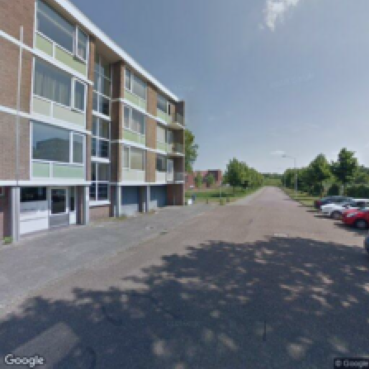}
        \caption*{}
    \end{subfigure}

    \caption*{\textbf{high, very high risk:}}
    \begin{subfigure}[b]{0.24\textwidth}
        \includegraphics[width=\textwidth]{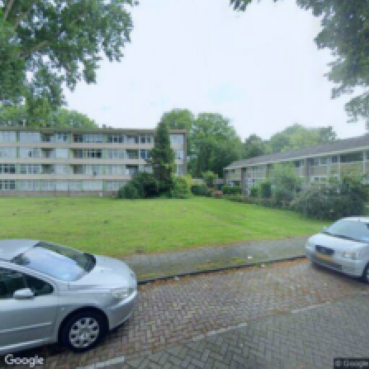}
        \caption*{}
    \end{subfigure}
    \begin{subfigure}[b]{0.24\textwidth}
        \includegraphics[width=\textwidth]{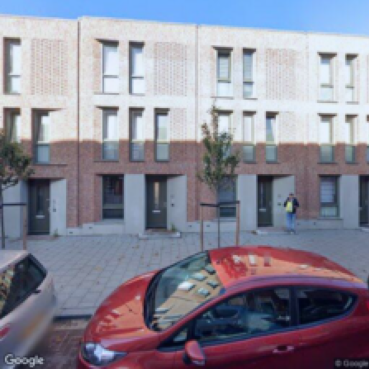}
        \caption*{}
    \end{subfigure}
    \begin{subfigure}[b]{0.24\textwidth}
        \includegraphics[width=\textwidth]{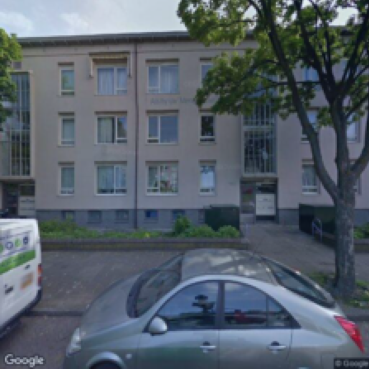}
        \caption*{}
    \end{subfigure}
    \begin{subfigure}[b]{0.24\textwidth}
        \includegraphics[width=\textwidth]{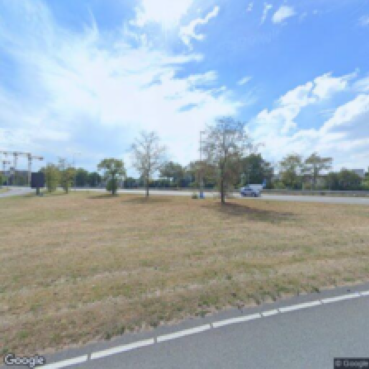}
        \caption*{}
    \end{subfigure}
    
    \caption{Street-view images that were correctly identified by Deit Base as pointing to the given level of depression and anxiety risk, per level of the target variable. Images: Google Street View.}
    \label{fig:fig12}
\end{figure}

An examination of high-confidence true positive predictions revealed distinct features for each class (see Figure \ref{fig:fig12}):

\begin{itemize}
\item ``Very low risk'' images typically displayed fields and expansive sky areas without any built structures, mirroring ResNet's predictions.
\item ``Low risk'' images predominantly showcased single-family homes and duplexes, a departure from ResNet's usual portrayal. However, like in the case of the ResNet, these images often included curated green spaces.
\item ``Moderate risk'' images were characterized by apartment blocks, cars positioned farther away or absent altogether, and bicycles.
\item For the ``high, very high risk'' category, large apartment complexes dominated the imagery. 
\item Notably absent were terraced houses, which were not present in the top-confidence predictions for any class in this model.
\end{itemize}

SHAP analysis indicated that, similar to ResNet50, DeiT Base relied on sky areas but differed in the importance of cars and bicycles (see figures in the Supplementary Materials). These features, however, were ambiguous in their influence on classification, as evidenced by extreme SHAP values in both directions. 

The attention weight visualizations did not offer clear interpretative insights and are detailed in the Supplementary Materials.

\section{Discussion and Conclusion} \label{sec:disc-concl}

This study examined the potential of deep learning (DL) models, namely DeiT and ResNet, in predicting depression and anxiety risks using Dutch Street View Images (SVIs). This research provides initial evidence suggesting DL models, when trained on SVIs, might be effective in discerning neighborhood-level depression and anxiety risks. Further studies are essential to enhance the models' interpretability and to guide public health strategies on the significance of specific visual environmental factors in relation to depression and anxiety risks.\\

\textbf{Key findings.} Both the DeiT and ResNet models displayed a propensity to misclassify neighboring risk categories, likely due to the continuous nature of the depression and anxiety risk scores. Initial analyses of high-confidence, true-positive predictions highlighted class-distinct features, such as the balance between greenery and the built environment, a finding consistent with prior research \citep{helbich2019, wang2019}.

While earlier studies centered on extracting and assessing the predictive capability of features like greenery or sky views on mental health outcomes, the visualization of SHAP values for our fully fine-tuned ResNet model indicates the importance of expansive sky areas and tree canopies. Interestingly, these insights were derived directly from images without manually specifying features of interest. Further, distinctions such as the presence of cars versus expansive sky areas appeared critical for accurate classifications. Other elements like roofs, shaded walls, and shadows beneath roofs, though not highlighted in prior research, emerged as significant for accurate predictions. Notably, roofs and their associated shadows seem as vital as the general greenery and nearly as influential as the specific tree canopy.

These patterns hint at a potential correlation between specific visual elements in a neighborhood and the risk of depression. This observation aligns with prior research that linked visual elements like water, sky, and greenery with mental well-being \citep{helbich2019, wang2019}. This connection is further underscored by our examination of high-confidence true-positive predictions (see Figure \ref{fig:fig12}). Here, each risk category appeared to possess a unique balance of greenery versus built environment, mirroring some findings from past studies \citep{wang2019, helbich2019, ki2021}. However, earlier investigations mostly zoomed in on greenery, sky, and water indices without juxtaposing them against the built environment ratio. Our results suggest variations in the types of built environments related to each risk level, pointing to a need for more in-depth exploration in literature to understand the relationship between built environment types, greenery, and mental health risks.\\

\textbf{Limitations and Challenges.} While both the ResNet50 and DeiT networks performed well on the classification task, their accuracy remains far from optimal. This may be due not only to the limited amount of data used for training, but also because of the multitude of factors contributing to depression and anxiety risk \citep{kohler2018}, many of which may not be visible in SVIs. For example, many known factors are due to genetics, diet, infections, or trauma, which would not be evident from street-view images. At the same time, as \cite{keralis2020} previously pointed out, there are factors that are environmental in nature, but are not represented in the SVI, for example perceived safety and traffic congestion.

The 'high depression and anxiety risk' variable in this study was sourced from Dutch Health Monitor statistics, which rely on Kessler's questionnaire rather than Diagnostic and Statistical Manual of Mental Disorders (DSM) criteria for their categorizations (see Methods for more details). This was a deliberate choice, as access to clinically diagnosed cases disaggregated by zip codes would entail handling sensitive personal data. Given the exploratory nature of this pilot study, utilizing such data was deemed ethically unsuitable.

The study cannot suit a purpose of causal inference about the reasons for depression and anxiety disorders' prevalence as it doesn't capture temporal precedence and doesn't exclude any confounding variables.\\

\textbf{Final Remarks.} This study provides preliminary evidence for the utility of deep learning to aid the monitoring of environmental factors correlated with mental health wellness. Future work should focus on improving the efficacy of Explainable AI techniques in this domain, and on using them to assess the importance of specific environment features on the risk of depression and anxiety. Similar studies can suggest environmental factors strongly correlated with depression and anxiety disorders, that could then be further investigated with methods allowing for causal inference. We finally suggest that applications similar to those developed in this work may be useful for real-time monitoring of the impact of changes in the environment over time on mental health. This could serve as a simpler alternative to the biennial Dutch Health Monitor, allowing for more frequent assessments of neighborhood mental health risks.

\subsubsection{Acknowledgements.} We extend our gratitude to Dr. Murat Kirtay for his insightful feedback and advice.

\bibliography{references}

\section{Supplementary Materials}
\label{sec:suppl}

\subsection{Supplementary Methods}
\subsubsection{Dataset}

\begin{figure}[h]
    \centering
    \begin{subfigure}{.5\textwidth}
        \centering
        \includegraphics[width=.9\linewidth]{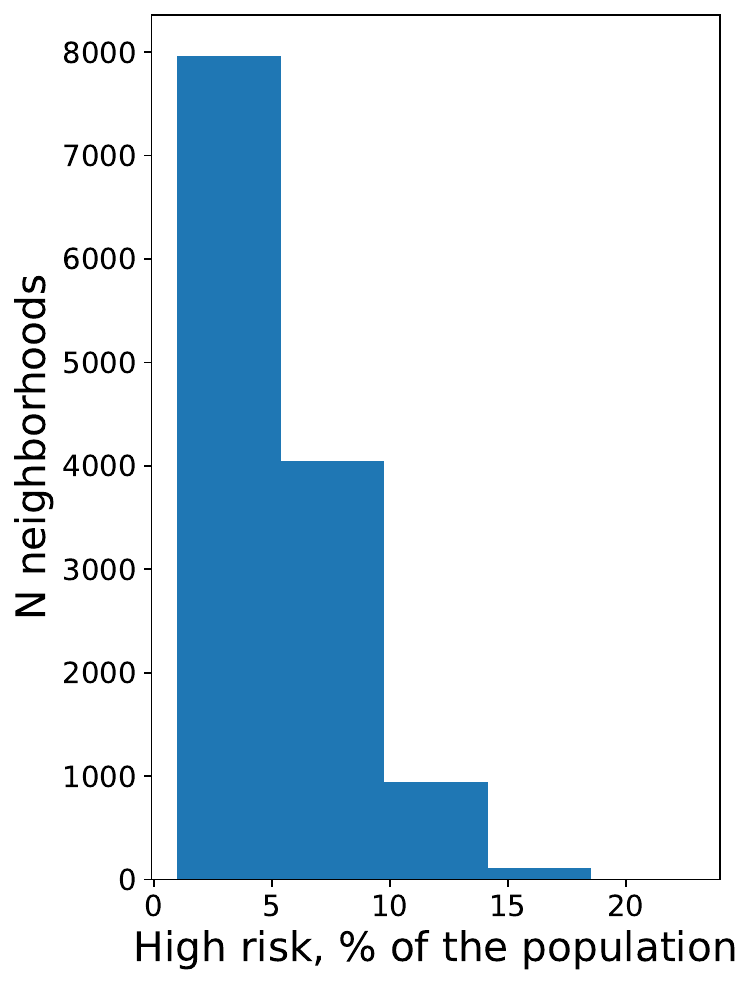}
        \caption{}
        \label{fig:sub1-1}
    \end{subfigure}%
    \begin{subfigure}{.5\textwidth}
        \centering
        \includegraphics[width=.9\linewidth]{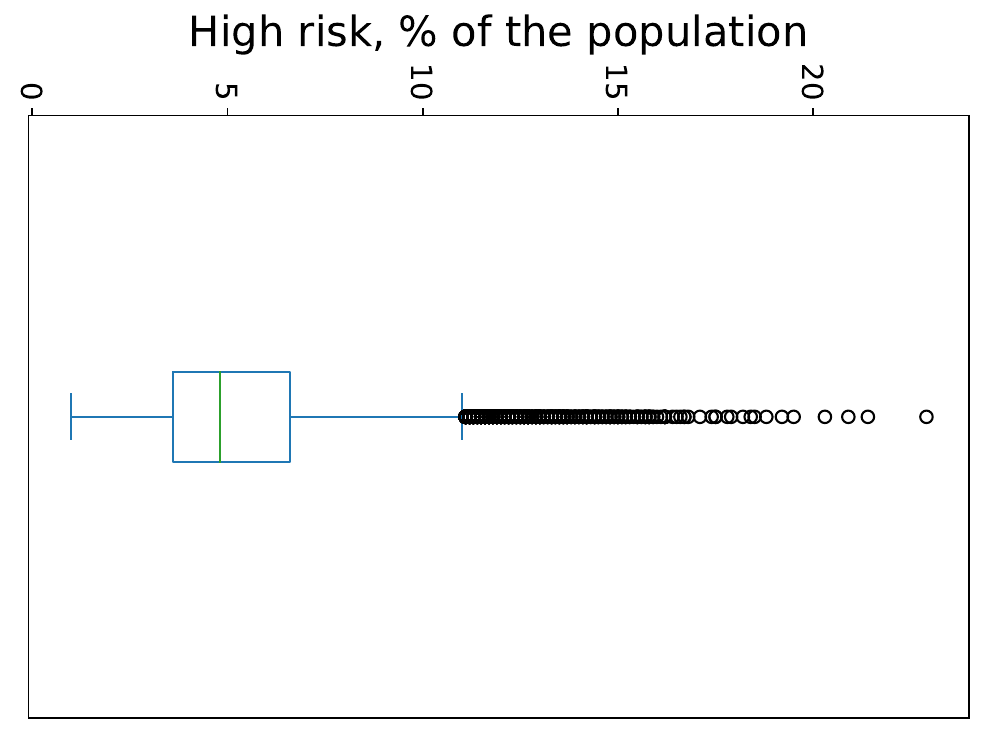}
        \caption{}
        \label{fig:sub2-1}
    \end{subfigure}
    \caption{Distribution of the target variable. Left: the histogram of the number of neighbourhoods with respect to the depression and anxiety risk scores. The histogram is split into 5 bins in order to represent the distribution of the target variable’s classes. Right: box plot showing the median, the quartiles and the spread of outliers of the depression and anxiety risk values.}
    \label{fig:test}
\end{figure}

\textbf{Original distribution of the target variable.} Figure \ref{fig:test} provides more information on how the original risk of depression and anxiety variable is distributed in the Dutch Health Monitor (\cite{gezondheidsmonitor2020}). 

\textbf{Details of the sampling procedure.} The Dutch Health Monitor’s data augmented with the discretized target variable was merged on neighbourhood codes with geographical data of \cite{kadaster2022}. The coordinates of the polygons were converted from the European Petroleum Survey Group's (EPSG) coordinate system 28992 to EPSG 4326. This conversion was done to transform the coordinates into the latitude-longitude format, which aligns with the way SVI data is retrieved. 

In order to create the input feature datasets, SVI of Google Street View (GSV) was sampled with Google Maps API. Only images that belong to Google were sampled. This ensures that no indoor images made by third parties are included in the dataset. Mapillary or KartaView could have been an alternative source of images. However, their coverage in many neighbourhoods in the Netherlands is limited to big streets and highways, which won’t be representative of the environment in which people live and walk. Also all their content is user-generated, which prevents from filtering images made indoors. Therefore, GSV was chosen as the source of images. The official GSV API was used for downloading, no scraping was used in order to comply with the terms of usage.  

Based on the counts of neighbourhoods per category (Table \ref{tab:intervals}), and with the aim to get around 10,000 images, the ‘very low’ and ‘low’ levels of depression and anxiety risk were represented with 2500 random neighbourhoods each, 1 image per neighbourhood, totalling 5,000 images. For the other categories it was not possible to sample one image per neighbourhood due to the limited amount of neighbourhoods. The ‘moderate’ class was represented by 3 images per neighbourhood for 833 randomly selected neighbourhoods, totalling 2499 images. After a preliminary exploration the ‘very high’ category was added to the ‘high’ category, as 7 neighbourhoods were too little to provide 2,000 images on their own without causing overfitting, and the choice was made to merge the two categories and preserve the balance of the dataset, as both categories represent outliers anyways. For the neighbourhoods of this fourth merged category, 20 images per neighbourhood were sampled, totalling 2,380 images. 

While sampling, 70 neighbourhoods returned no points for the first level within the given timeframe (May-September) after 300 attempts per neighbourhood. In order to replace these neighbourhoods, another 70 neighbourhoods of the same class were sampled randomly. After resampling another 3 neighbourhoods returned no points and another resampling was done for these neighbourhoods. For the second level, 26 neighbourhoods were resampled in one go, for the third level 8 neighbourhoods had to be resampled, then 1 had to be resampled. In the fourth level all the neighbourhoods had 20 points after the first go. 

\begin{figure}[h]
    \centering
    \includegraphics[width=0.5\textwidth]{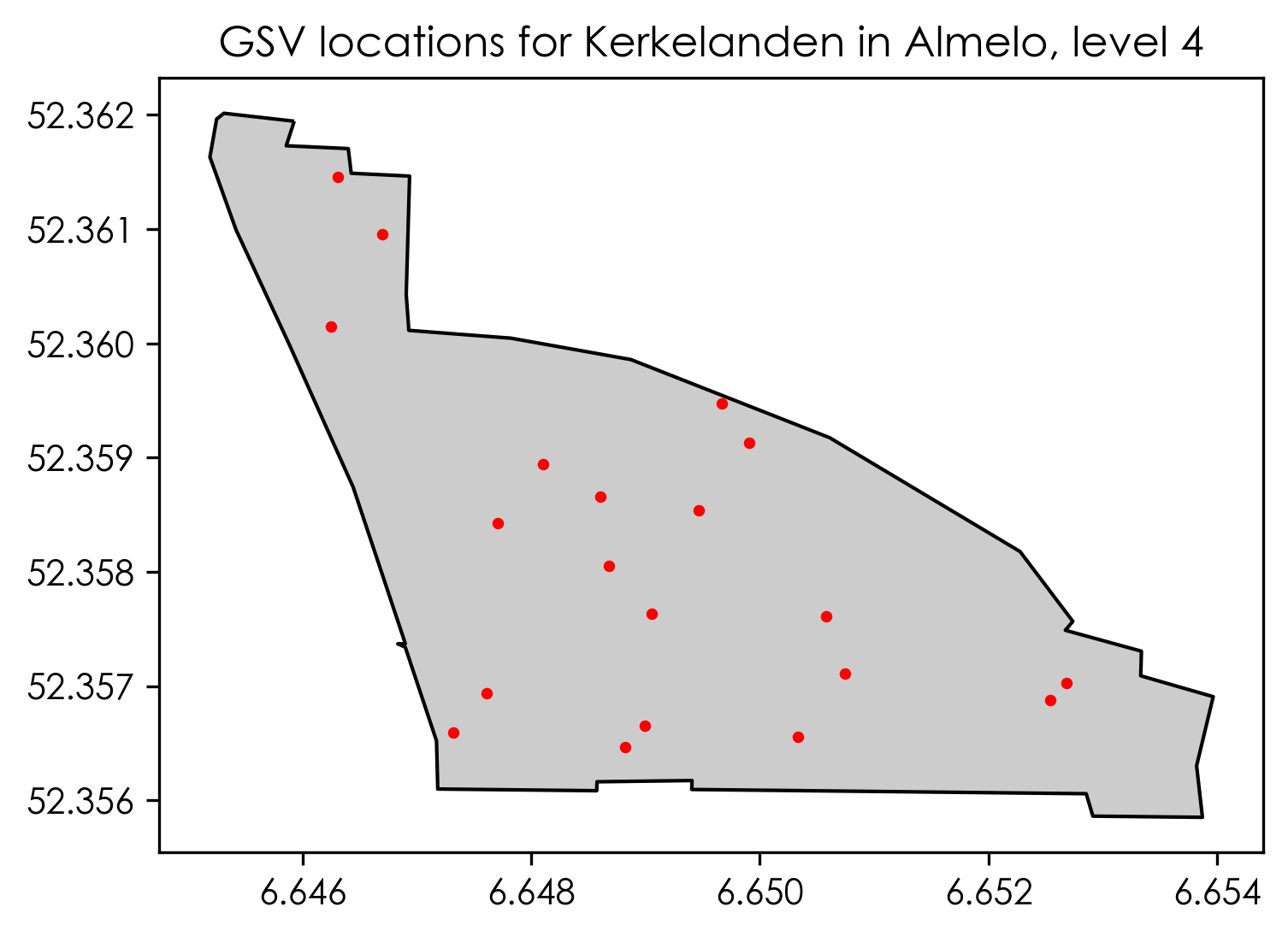} 
    \caption{Typical non-biased sampling of coordinates within a neighbourhood in the final dataset.}
    \label{fig:fig7}
\end{figure}

\begin{figure}[h]
    \centering
    \begin{subfigure}{.45\textwidth}
        \centering
        \includegraphics[width=.9\linewidth]{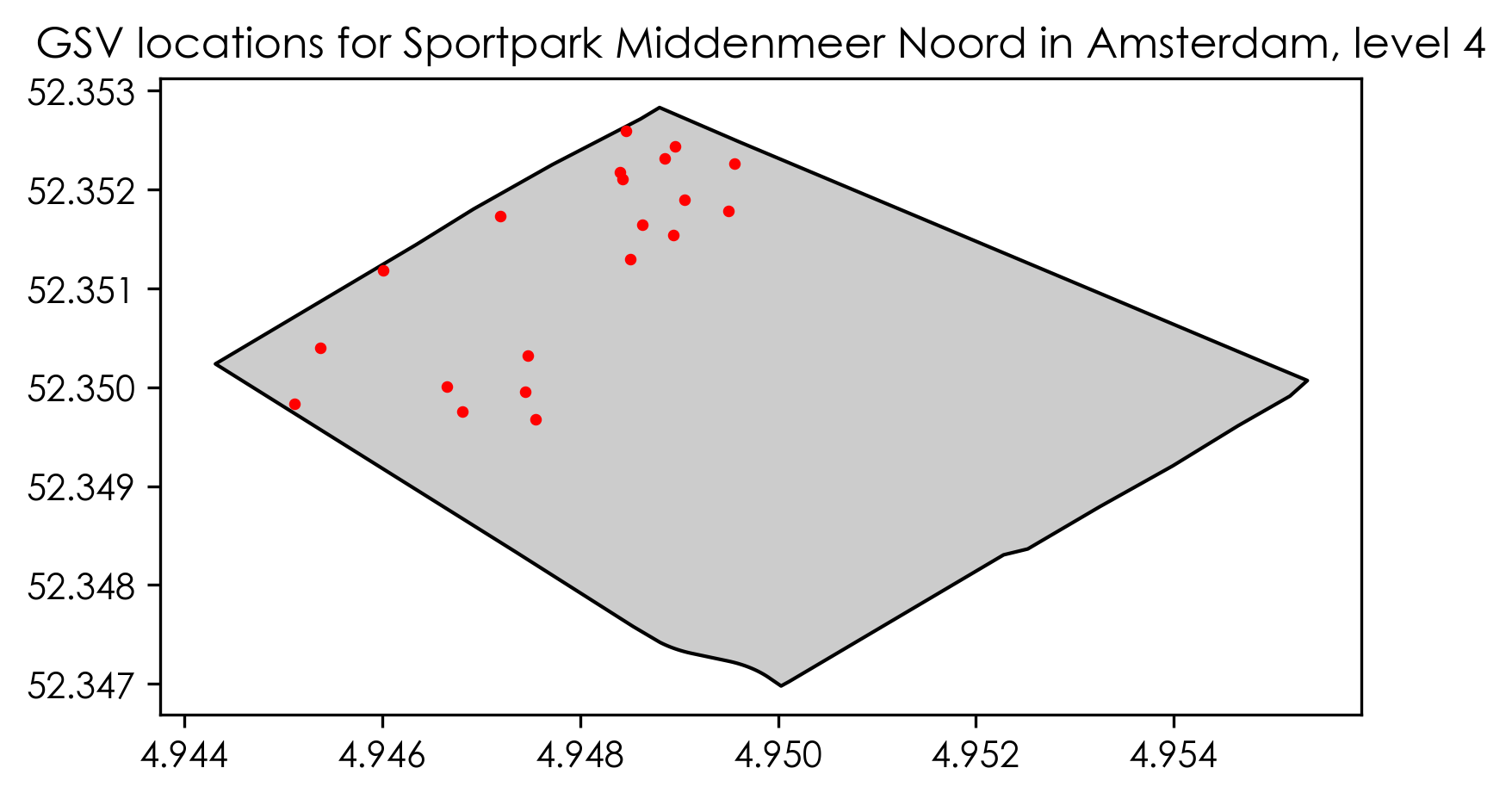}
        \caption{}
        \label{fig:sub1}
    \end{subfigure}%
    \begin{subfigure}{.45\textwidth}
        \centering
        \includegraphics[width=.9\linewidth]{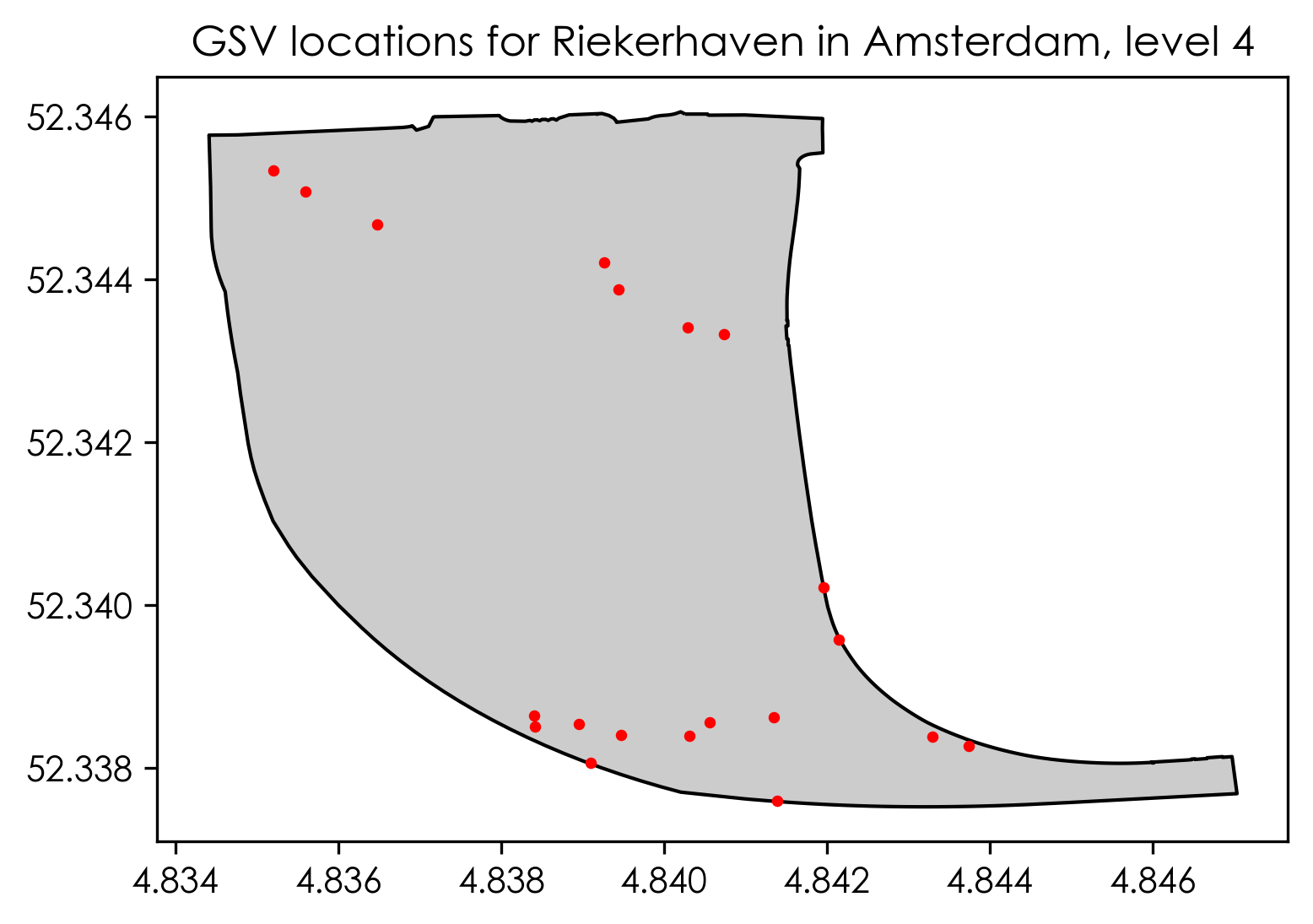}
        \caption{}
        \label{fig:sub2}
    \end{subfigure}
    \newline % Creates a new line for the next set of subfigures
    \begin{subfigure}{.45\textwidth}
        \centering
        \includegraphics[width=.9\linewidth]{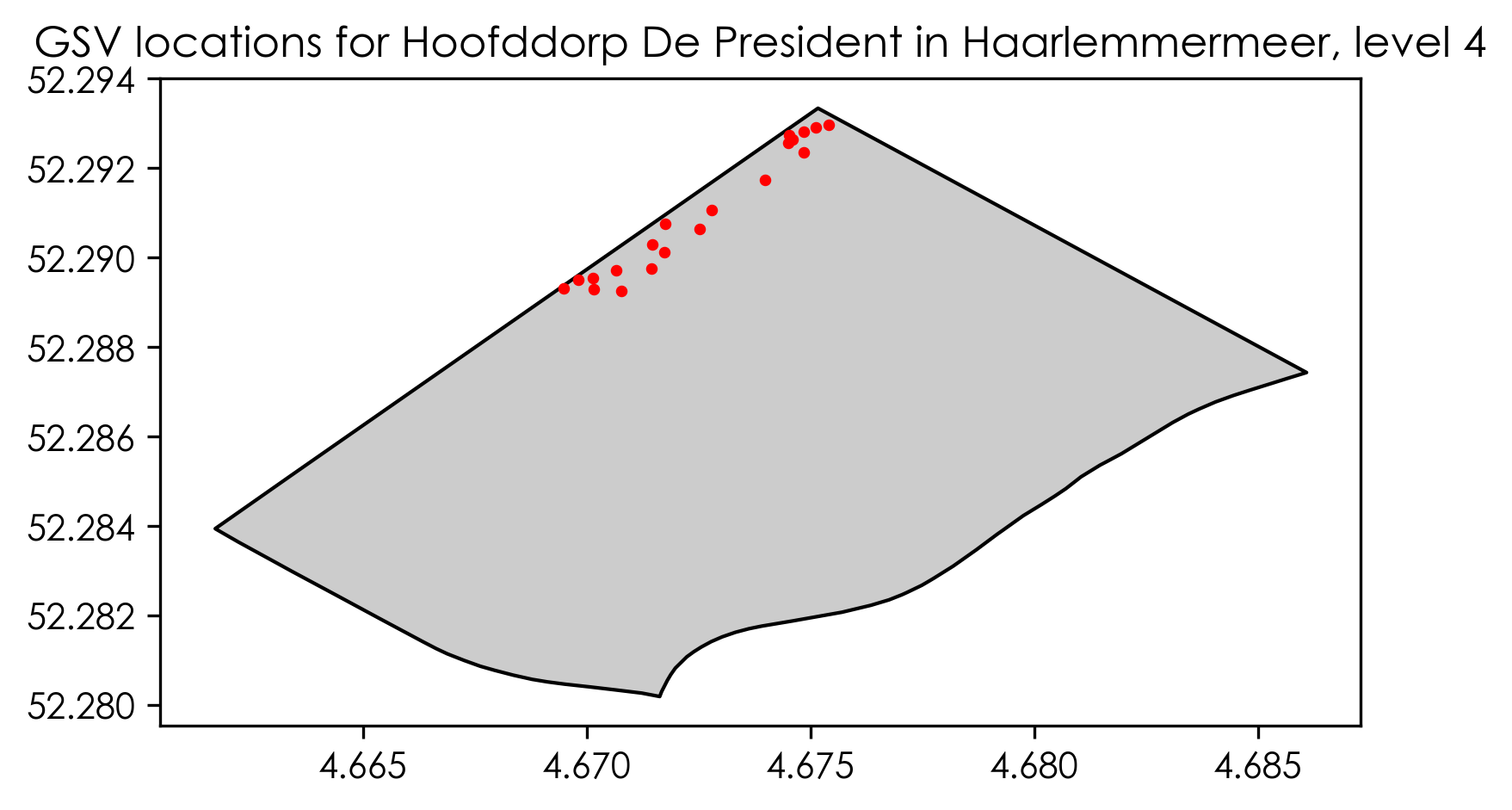}
        \caption{}
        \label{fig:sub3}
    \end{subfigure}%
    \begin{subfigure}{.45\textwidth}
        \centering
        \includegraphics[width=.9\linewidth]{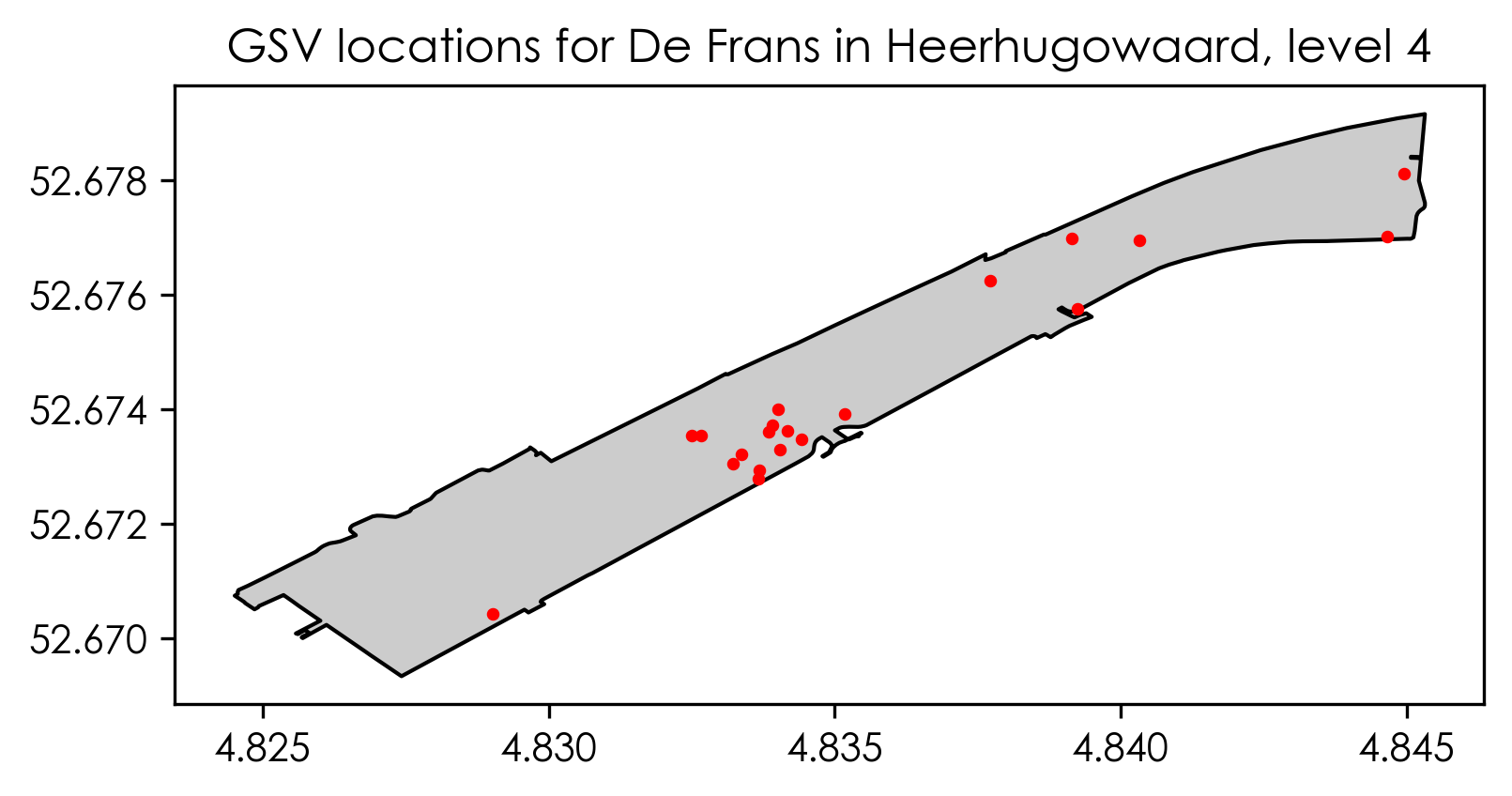}
        \caption{}
        \label{fig:sub4}
    \end{subfigure}
    \caption{Examples of sampling that overrepresents certain parts of the neighourhoods to a different extent.}
    \label{fig:fig8}
\end{figure}

The united “High and very high” level category had the most images per neighbourhood. In order to control for the places-overrepresentation bias, coordinates of all the images per neighbourhood were visualized. Based on the visualizations, it seems like 20 points per neighbourhood allowed for unbiased dispersion of the points for the most neighbourhoods, as in Figure \ref{fig:fig7}. Nevertheless, 16 out of 119 neighbourhoods exhibited some level of biased sampling, with points centered around certain parts of the neighbourhood, like for the neighbourhoods in Figure \ref{fig:fig8}. 

\textbf{Image characteristics.} GSV API allows for control of the camera facing of the given image by specifying 0, 90, 180, and 270-degree headings. However, for the datasets curated for this study the heading was not specified, and there was only one image collected per coordinates pair instead of 4 images (one per heading). Sampling GSVI with these specific headings could potentially be beneficial for constructing panoramas. However, given that pre-trained networks typically accommodate standard image shapes, predominantly square, the feasibility of feeding panoramas into these networks is limited. Therefore, a decision was made to compile a set of randomly selected square images for this study. In the meanwhile, the headings of GSVI always point in the direction of North by default, they don’t align with the orientation of the road or any other specific features. Consequently, the resulting images might be as random as a simple collection of arbitrary images from the area.

It is also possible to control the field of view for GSVI download, making it bigger or smaller. This way the images could include more of the scenery. However, bigger camera angle results in distorted images. \cite{ki2021} warn about it: they had to crop their panorama’s in order to get rid of all the distorted parts, otherwise segmentation misclassified too much. Therefore, the datasets utilized for this study use the default field of view, that proves to not distort the data.

\begin{figure}[htbp]
    \centering

    \begin{tabularx}{\textwidth}{XXXX}
    Original & Mean, discard 0\% & Max, discard 90\% \\
    \end{tabularx}

    \begin{subfigure}[b]{.3\textwidth}
        \centering
        \includegraphics[width=.9\linewidth]{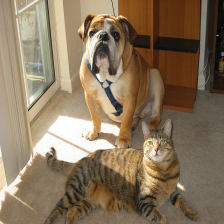}
        \caption*{from \cite{gildenblat2020}}
        \label{fig:g1}
    \end{subfigure}%
    \begin{subfigure}[b]{.3\textwidth}
        \centering
        \includegraphics[width=.9\linewidth]{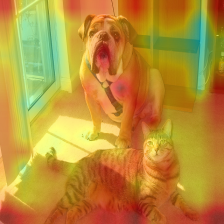}
        \caption*{}
        \label{fig:g2}
    \end{subfigure}
    \begin{subfigure}[b]{.3\textwidth}
        \centering
        \includegraphics[width=.9\linewidth]{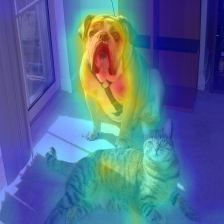}
        \caption*{}
        \label{fig:g3}
    \end{subfigure}%

    \begin{subfigure}[b]{.3\textwidth}
        \centering
        \includegraphics[width=.9\linewidth]{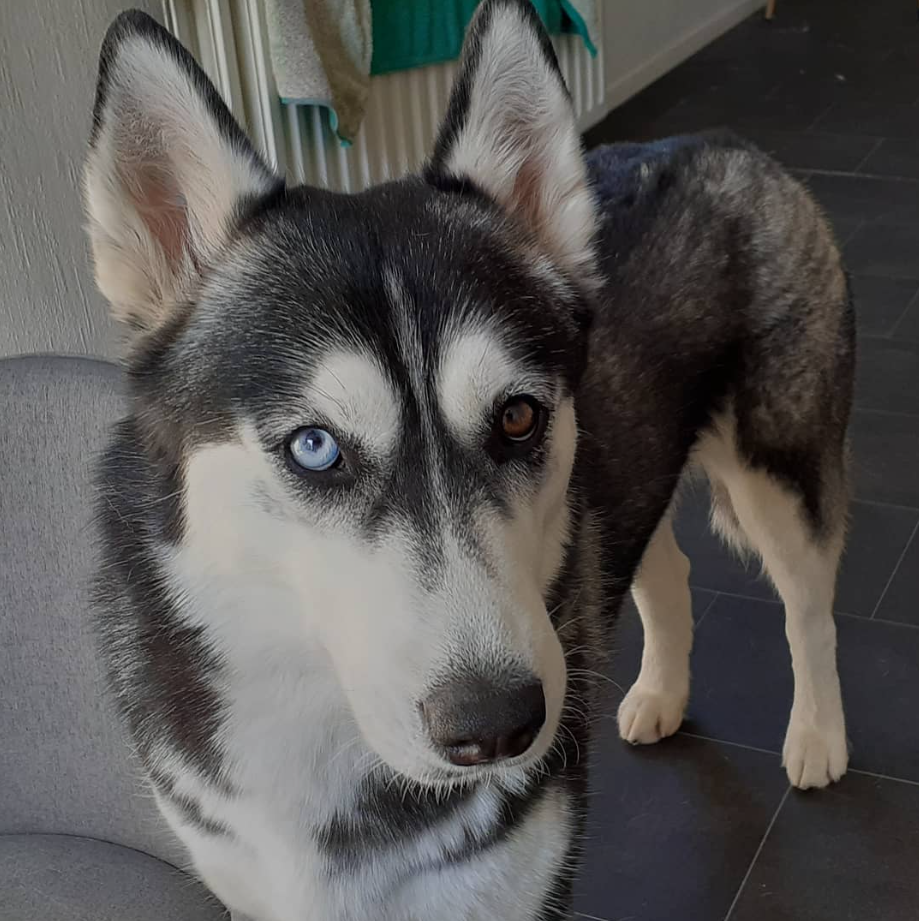}
        \caption*{our own image}
        \label{fig:g1}
    \end{subfigure}%
    \begin{subfigure}[b]{.3\textwidth}
        \centering
        \includegraphics[width=.9\linewidth]{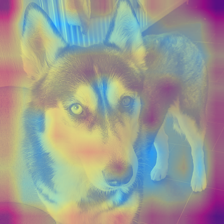}
        \caption*{}
        \label{fig:g2}
    \end{subfigure}
    \begin{subfigure}[b]{.3\textwidth}
        \centering
        \includegraphics[width=.9\linewidth]{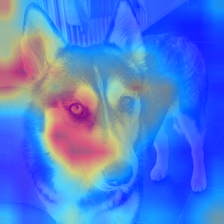}
        \caption*{}
        \label{fig:g3}
    \end{subfigure}%

    \caption{Illustration of the Gildenblat's pipeline \citep{gildenblat2020}. First row presents the author's example, re-run through the pipeline for this study. Second row presents our own image run through the pipeline in order to verify it works as intended on the kind of images it was originally designed for. First column - the image itself. Second column - image with average attentions vizualized, not informative to a human, according to the author. Third column - image with discard ratio for the attention map set to 90\% with only 10\% of the maximal weights visualized: the attention map becomes more crisp and informative.}
    \label{fig:gild}
\end{figure}

\textbf{Parameters of Attention Rollout.} For DeiT model, Gildenblat’s adaptation of attention rollout (2020) was used with minor adaptation. The pipeline designed by Gildenblat not only offers visualization of attention weights, but it also allows for the selective display of extreme weights, providing a clearer, more interpretable view of the model's decision-making process. It presents the possibility to choose the way of attention heads' fusion between mean, max, or min. The pipeline also incorporates a discard rate to enhance the crispness and interpretability of the visualizations (Figure \ref{fig:gild}). Discard ratio was set at 0.8, which meant that 80\% of the least extreme parameters were discarded prior to visualization. VITAttentionRollout function from vit-explain package was used to produce the visualizations, with minor adaptations in order to ensure the fine-tuned model is used and the images are being saved as separate files. 

\subsection{Hyperparameter Selection Results}

The best set of hyperparameters found during hyperparameter search are included in Tables \ref{tab:hyperparameter1} and \ref{tab:hyperparameter2}.

\begin{table}[h]
    \centering
    \small
    \caption{Results of the hyperparameter tuning: DeiT family of models}
    \label{tab:hyperparameter1}
    \begin{tabular}{l l r r r}
        \toprule
        Model & Layers unfrozen & Optimizer & Learning rate & Val. acc. change\\
        &&&&in 15 epochs, \\ &&&&percentage points\\
        \midrule
        Deit Base  & 1 & Adam & 0.005 & 12.87 \\
        Deit Small & 5 & SGD & 0.001 & 11.96 \\
        Deit Small & 3 & Adagrad & 0.005 & 11.85 \\
        Deit Tiny & 5 & SGD & 0.001 & 10.64 \\
        Deit Tiny & 1 & SGD & 0.001 & 10.53 \\
        \bottomrule
    \end{tabular}
\end{table}

\begin{table}[h]
    \centering
    \small
    \caption{Results of the hyperparameter tuning: ResNet50}
    \label{tab:hyperparameter2}
    \begin{tabular}{l l r r r}
        \toprule
        Model & Layers unfrozen & Optimizer & Learning rate & Val. acc. change\\
        &&&&in 15 epochs, \\ &&&&percentage points\\
        \midrule
        ResNet50 & 3 & Adagrad & 0.01 & 10.33 \\
        ResNet50 & 1 & SGD & 0001 & 8.81 \\
        ResNet50 & 5 & RMSprop & 0.001 & 8.41 \\
        ResNet50 & 5 & RMSprop & 0.005 & 7.90 \\
        ResNet50 & 5 & Adam & 0.01 & 7.60 \\
        ResNet50 & 3 & RMSprop & 0.01 & 7.60 \\
        
        \bottomrule
    \end{tabular}
\end{table}

\subsection{Results: overview of the types of landscapes per class}

\subsubsection{ResNet50.} Exploration of the true positive predictions with the highest logits revealed that for top-certainty true positive predictions, the depicted landscapes in the provided examples share analogous properties within their respective classes (Figure \ref{fig:best_resnet}). 

Specifically, in the "very low risk" class, the images that the model accurately classified with the highest certainty predominantly showcase rural landscapes; among the top 10 high-certainty images, 7 feature an absence of built environments, while the remaining pictures display single-family homes. These images exhibit a lot of greenery and sky in these images. 

Within the "low risk" class, images mostly feature terraced houses with exception for two out of 10 being duplex houses and one being a single-family home. A lot of sky and small cultivated greenery is visible on every image for the top-certainty correct predictions for this class. 

For both "moderate" and "high and very high" levels the buildings are mostly blocks of flats, cars are present at the images, along with little greenery and little sky. There are almost no cars for the "very low risk" level and always cars for the other levels. 

\subsubsection{DeiT Base.} For this model the images behind the true positive predictions with the highest logits also exhibit typical features per class, albeit the images are less homogeneous for the "high and very high" risk level (Figure \ref{fig:fig12}). 

In particular, model was very good at correctly classifying the images containing fields and vast sky areas with no built environment as belonging to the "very low risk" class. Nine out of 10 of the top-certainty correct predictions for this category are like that, with another one containing a farmhouse at a distance. 

The correctly classified examples of the "low risk" class predominantly contain images of small houses (single-family homes and duplexes) with cultivated greenery (well-rounded front yards), with one exception being the entrance to a shop and another exception being an image of two trees in front of unidentifiable type of houses, with trees occupying almost the whole image. 

The buildings in the "moderate risk" images are all blocks of flats. For this class the model correctly and certainly classified the images that typically contained no cars in the front (two out of 10 top-certainty predictions represent a car close to the camera), only at a distance or not at all, and also five out of 10 top-certainty images contained bicycles. 

At nine of the 10 top-certainty images of the "high, very high risk" class a block of flats occupied a significant portion of the image. 

Curiously, typical Dutch terraced houses are not present in the top-certainty predictions for this model, for any class. 

\subsection{SHAP visualizations}
\subsubsection{ResNet50.} 
\begin{figure}[htbp]
    \centering

    % Row for class "very low"
    \begin{subfigure}[b]{0.23\textwidth}
        \includegraphics[width=\textwidth]{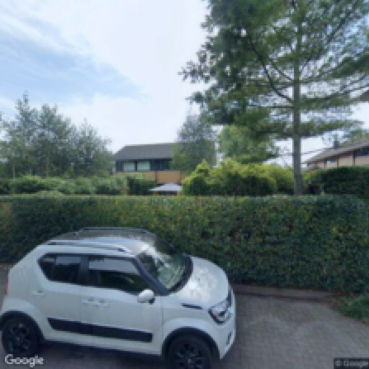}
        \caption*{}
    \end{subfigure}
    \begin{subfigure}[b]{0.23\textwidth}
        \includegraphics[width=\textwidth]{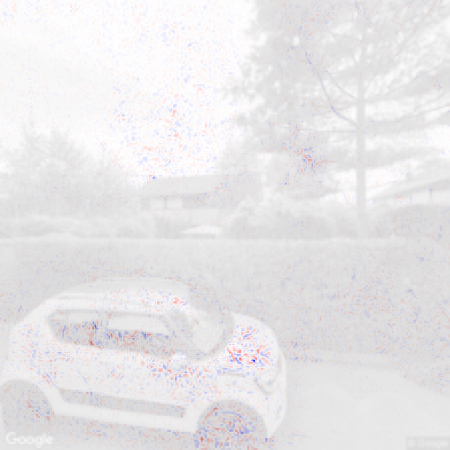}
        \caption*{}
    \end{subfigure}
    \begin{subfigure}[b]{0.23\textwidth}
        \includegraphics[width=\textwidth]{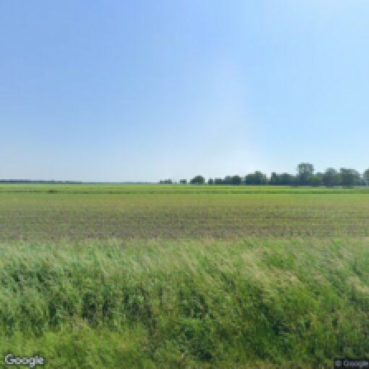}
        \caption*{}
    \end{subfigure}
    \begin{subfigure}[b]{0.23\textwidth}
        \includegraphics[width=\textwidth]{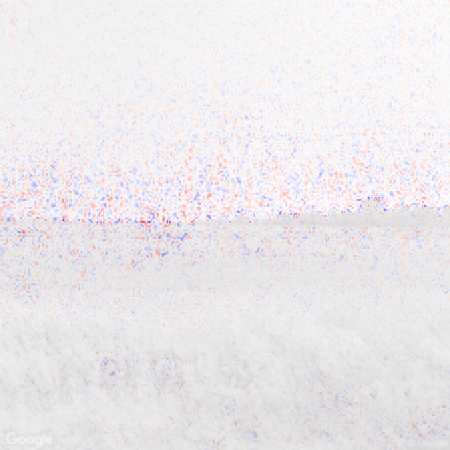}
        \caption*{}
    \end{subfigure}

        % Row for class "very low"
    \begin{subfigure}[b]{0.23\textwidth}
        \includegraphics[width=\textwidth]{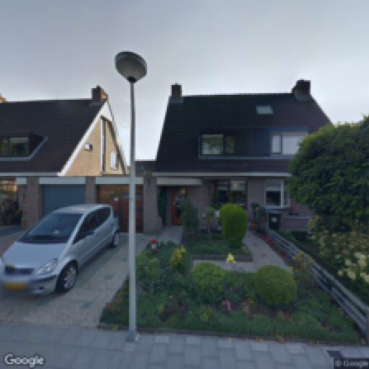}
        \caption*{}
    \end{subfigure}
    \begin{subfigure}[b]{0.23\textwidth}
        \includegraphics[width=\textwidth]{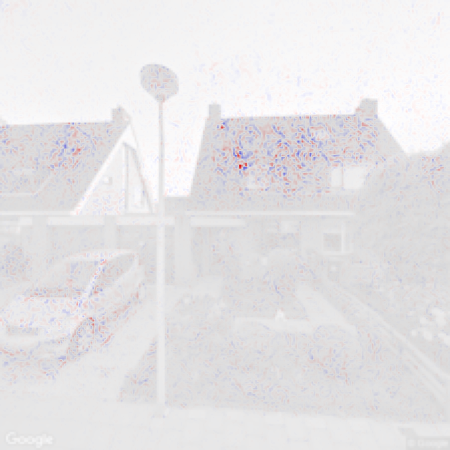}
        \caption*{}
    \end{subfigure}
    \begin{subfigure}[b]{0.23\textwidth}
        \includegraphics[width=\textwidth]{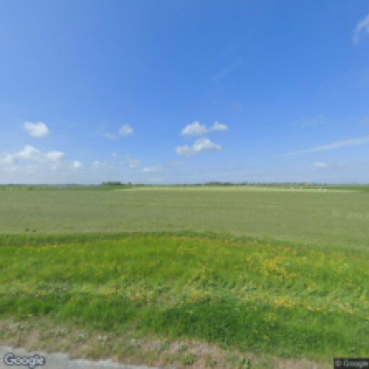}
        \caption*{}
    \end{subfigure}
    \begin{subfigure}[b]{0.23\textwidth}
        \includegraphics[width=\textwidth]{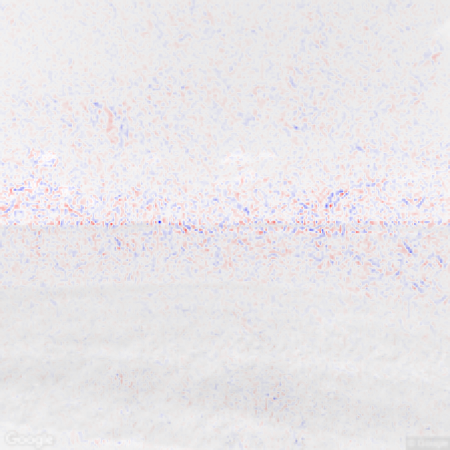}
        \caption*{}
    \end{subfigure}

        % Row for class "very low"
    \begin{subfigure}[b]{0.23\textwidth}
        \includegraphics[width=\textwidth]{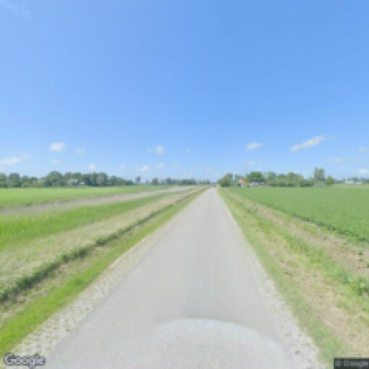}
        \caption*{very low risk}
    \end{subfigure}
    \begin{subfigure}[b]{0.23\textwidth}
        \includegraphics[width=\textwidth]{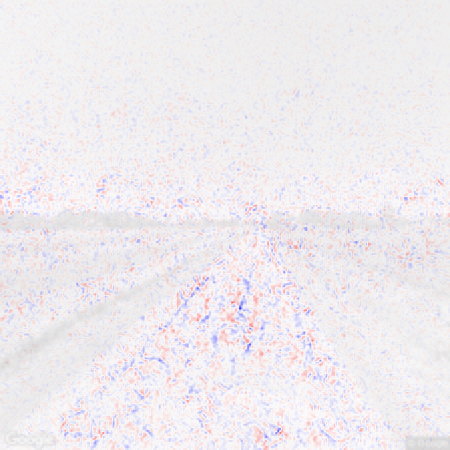}
        \caption*{}
    \end{subfigure}
    \begin{subfigure}[b]{0.23\textwidth}
        \includegraphics[width=\textwidth]{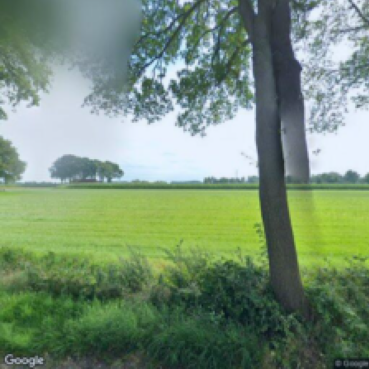}
        \caption*{}
    \end{subfigure}
    \begin{subfigure}[b]{0.23\textwidth}
        \includegraphics[width=\textwidth]{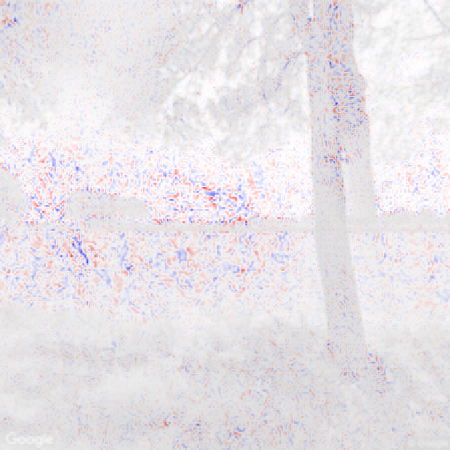}
        \caption*{}
    \end{subfigure}

       % Row for class "low"
    \begin{subfigure}[b]{0.23\textwidth}
        \includegraphics[width=\textwidth]{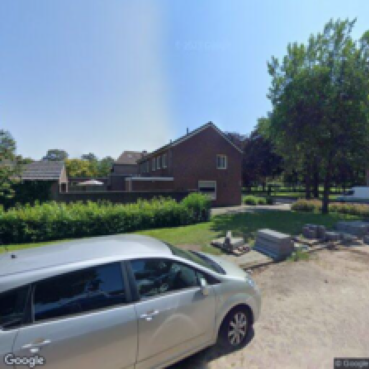}
        \caption*{}
    \end{subfigure}
    \begin{subfigure}[b]{0.23\textwidth}
        \includegraphics[width=\textwidth]{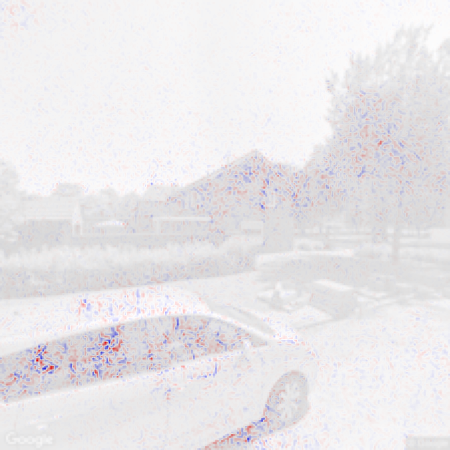}
        \caption*{}
    \end{subfigure}
    \begin{subfigure}[b]{0.23\textwidth}
        \includegraphics[width=\textwidth]{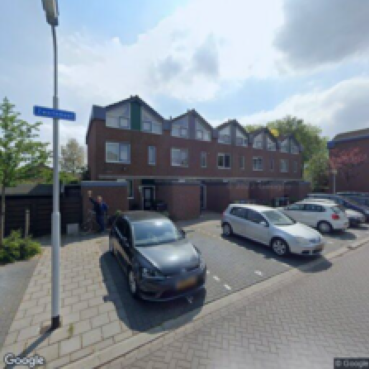}
        \caption*{}
    \end{subfigure}
    \begin{subfigure}[b]{0.23\textwidth}
        \includegraphics[width=\textwidth]{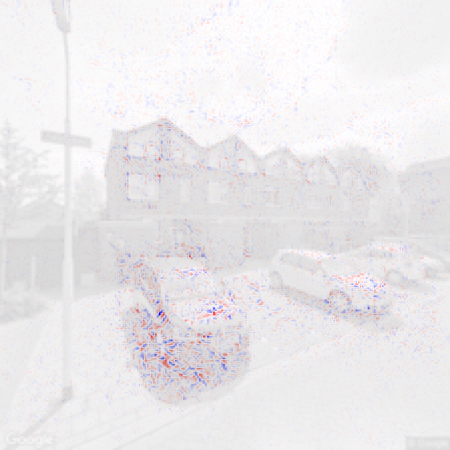}
        \caption*{}
    \end{subfigure}

        % Row for class "low"
    \begin{subfigure}[b]{0.23\textwidth}
        \includegraphics[width=\textwidth]{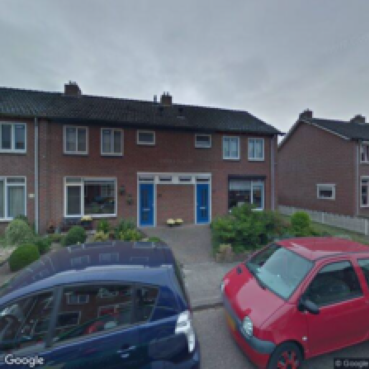}
        \caption*{}
    \end{subfigure}
    \begin{subfigure}[b]{0.23\textwidth}
        \includegraphics[width=\textwidth]{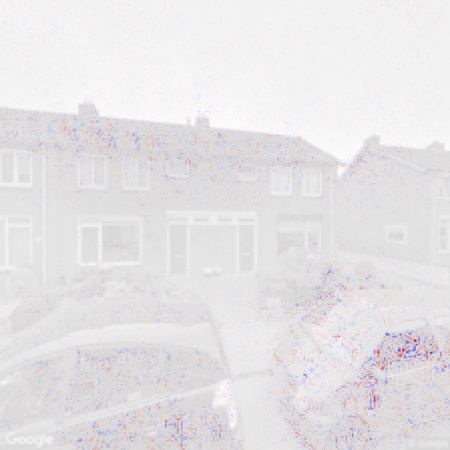}
        \caption*{}
    \end{subfigure}
    \begin{subfigure}[b]{0.23\textwidth}
        \includegraphics[width=\textwidth]{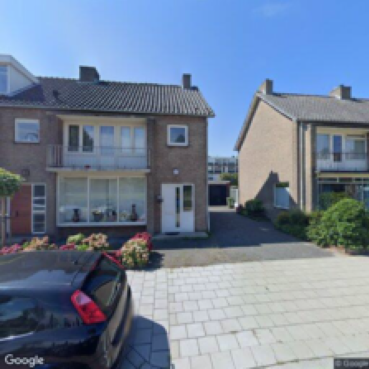}
        \caption*{}
    \end{subfigure}
    \begin{subfigure}[b]{0.23\textwidth}
        \includegraphics[width=\textwidth]{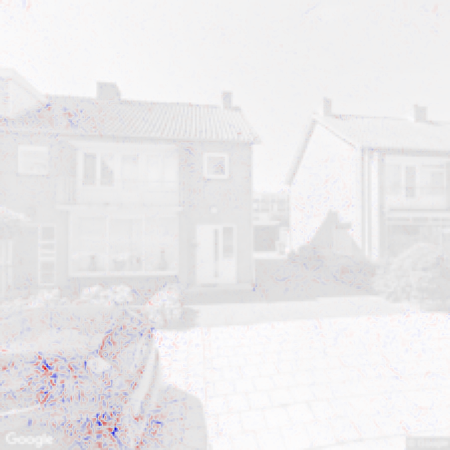}
        \caption*{}
    \end{subfigure}

        % Row for class "low"
    \begin{subfigure}[b]{0.23\textwidth}
        \includegraphics[width=\textwidth]{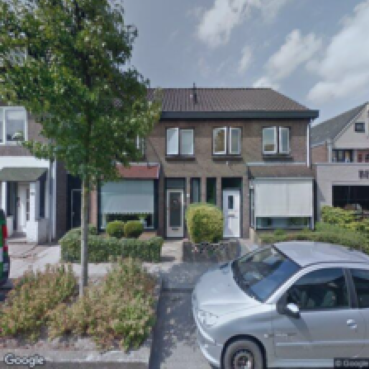}
        \caption*{low risk}
    \end{subfigure}
    \begin{subfigure}[b]{0.23\textwidth}
        \includegraphics[width=\textwidth]{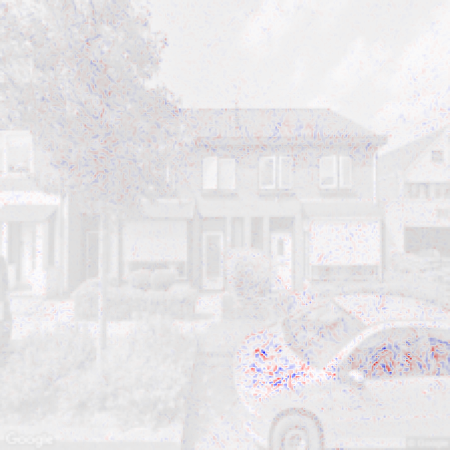}
        \caption*{}
    \end{subfigure}
    \begin{subfigure}[b]{0.23\textwidth}
        \includegraphics[width=\textwidth]{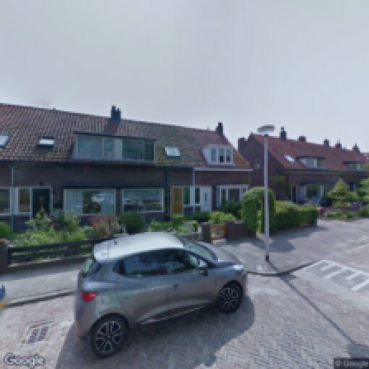}
        \caption*{}
    \end{subfigure}
    \begin{subfigure}[b]{0.23\textwidth}
        \includegraphics[width=\textwidth]{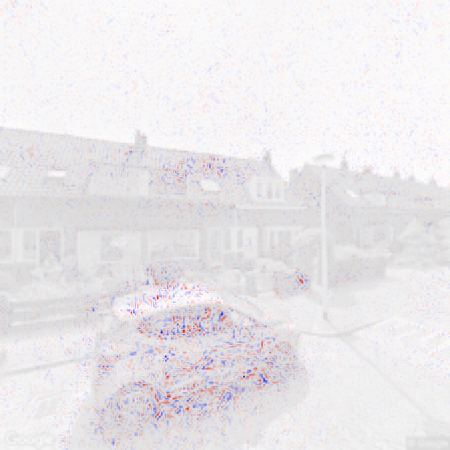}
        \caption*{}
    \end{subfigure}

    \caption{Visualization of SHAP values for the top-6 most certain predictions of ResNet50 per class. Part 1: very low and low risk. Images without overlay: Google Street View.}
    \label{fig:resnet_shap1}
\end{figure}

\begin{figure}[htbp]
    \centering

           % Row for class "moderate"
    \begin{subfigure}[b]{0.23\textwidth}
        \includegraphics[width=\textwidth]{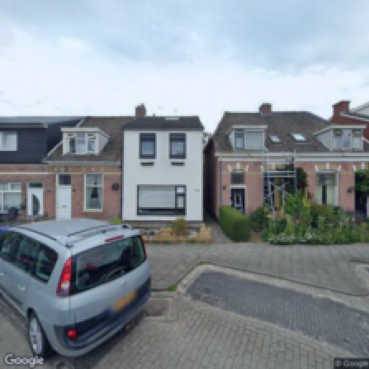}
        \caption*{}
    \end{subfigure}
    \begin{subfigure}[b]{0.23\textwidth}
        \includegraphics[width=\textwidth]{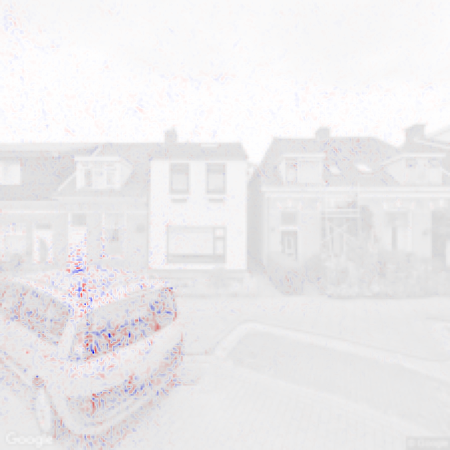}
        \caption*{}
    \end{subfigure}
    \begin{subfigure}[b]{0.23\textwidth}
        \includegraphics[width=\textwidth]{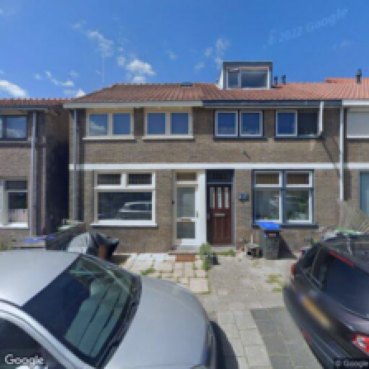}
        \caption*{}
    \end{subfigure}
    \begin{subfigure}[b]{0.23\textwidth}
        \includegraphics[width=\textwidth]{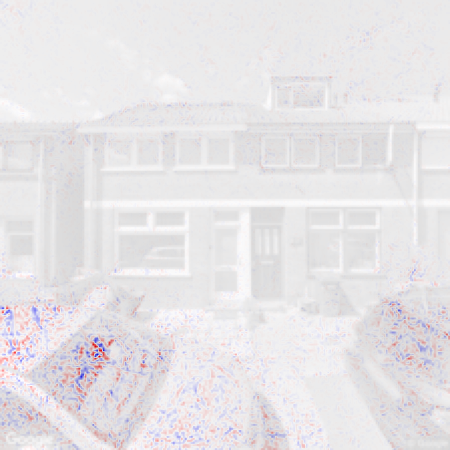}
        \caption*{}
    \end{subfigure}

        % Row for class "moderate"
    \begin{subfigure}[b]{0.23\textwidth}
        \includegraphics[width=\textwidth]{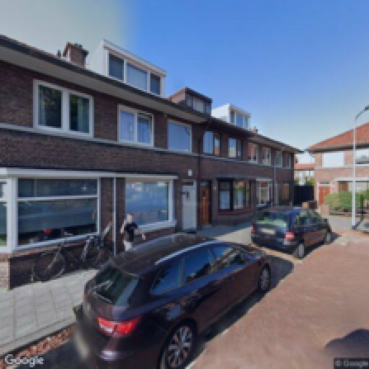}
        \caption*{}
    \end{subfigure}
    \begin{subfigure}[b]{0.23\textwidth}
        \includegraphics[width=\textwidth]{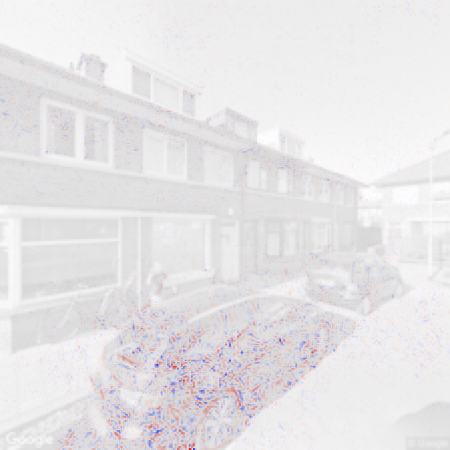}
        \caption*{}
    \end{subfigure}
    \begin{subfigure}[b]{0.23\textwidth}
        \includegraphics[width=\textwidth]{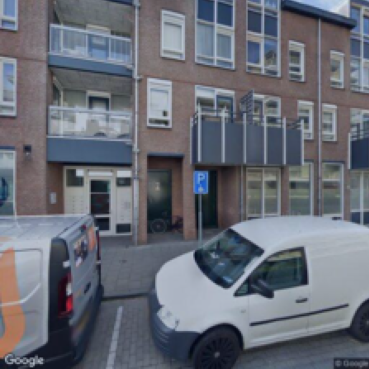}
        \caption*{}
    \end{subfigure}
    \begin{subfigure}[b]{0.23\textwidth}
        \includegraphics[width=\textwidth]{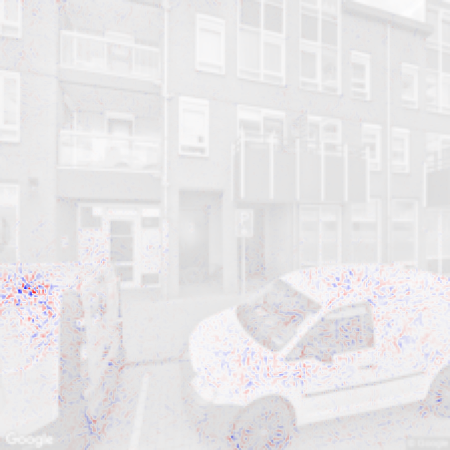}
        \caption*{}
    \end{subfigure}

        % Row for class "moderate"
    \begin{subfigure}[b]{0.23\textwidth}
        \includegraphics[width=\textwidth]{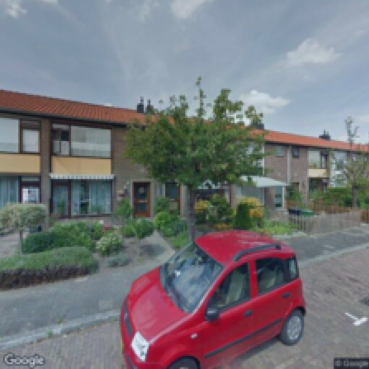}
        \caption*{moderate risk}
    \end{subfigure}
    \begin{subfigure}[b]{0.23\textwidth}
        \includegraphics[width=\textwidth]{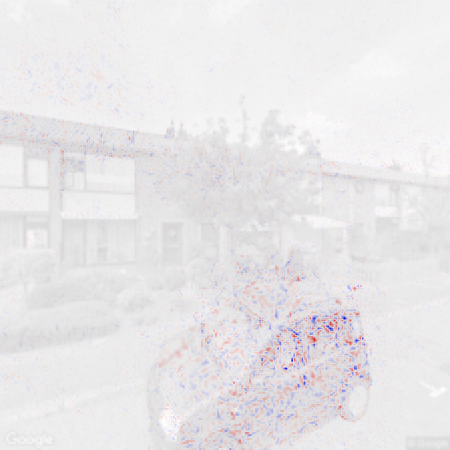}
        \caption*{}
    \end{subfigure}
    \begin{subfigure}[b]{0.23\textwidth}
        \includegraphics[width=\textwidth]{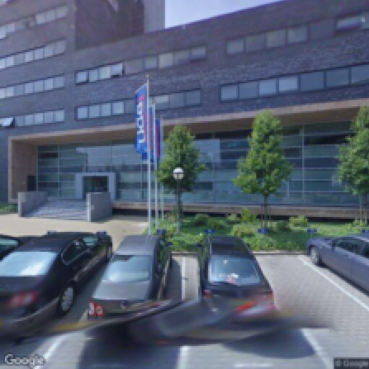}
        \caption*{}
    \end{subfigure}
    \begin{subfigure}[b]{0.23\textwidth}
        \includegraphics[width=\textwidth]{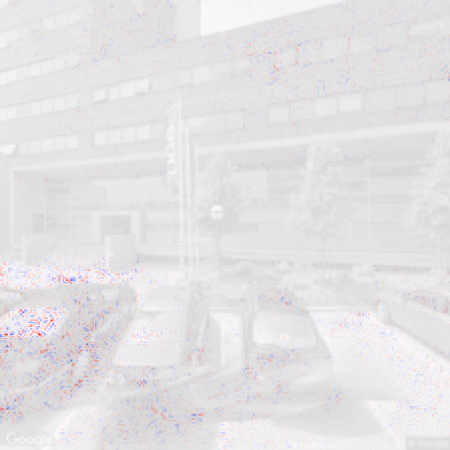}
        \caption*{}
    \end{subfigure}

           % Row for class "high and very high"
    \begin{subfigure}[b]{0.23\textwidth}
        \includegraphics[width=\textwidth]{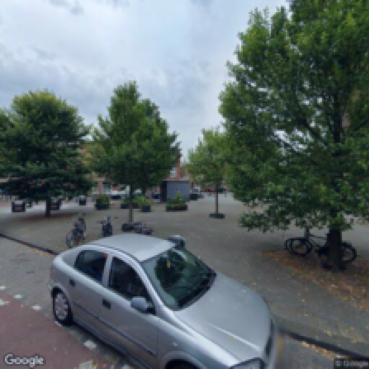}
        \caption*{}
    \end{subfigure}
    \begin{subfigure}[b]{0.23\textwidth}
        \includegraphics[width=\textwidth]{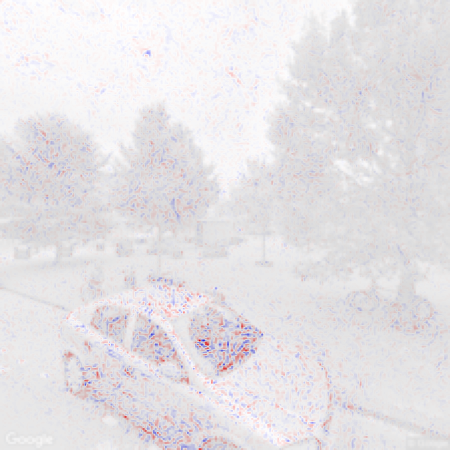}
        \caption*{}
    \end{subfigure}
    \begin{subfigure}[b]{0.23\textwidth}
        \includegraphics[width=\textwidth]{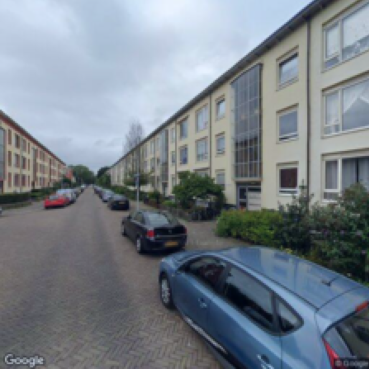}
        \caption*{}
    \end{subfigure}
    \begin{subfigure}[b]{0.23\textwidth}
        \includegraphics[width=\textwidth]{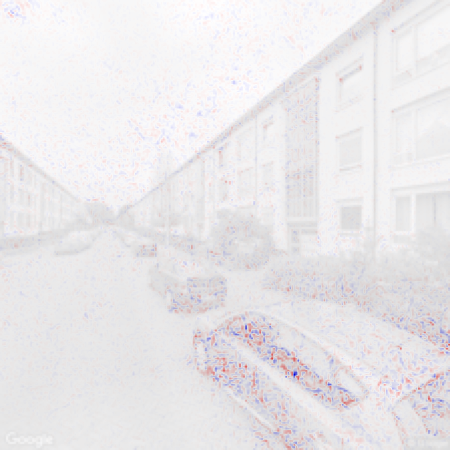}
        \caption*{}
    \end{subfigure}

        % Row for class "high and very high"
    \begin{subfigure}[b]{0.23\textwidth}
        \includegraphics[width=\textwidth]{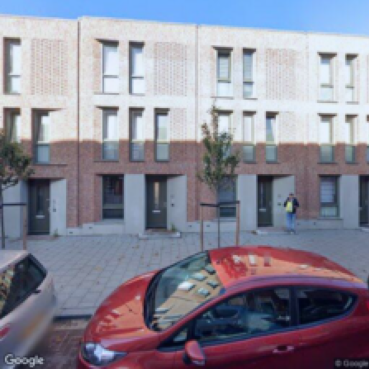}
        \caption*{}
    \end{subfigure}
    \begin{subfigure}[b]{0.23\textwidth}
        \includegraphics[width=\textwidth]{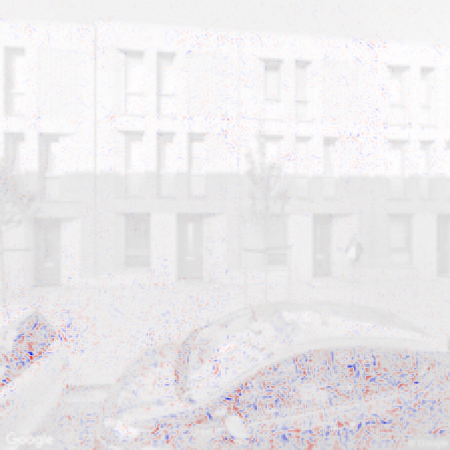}
        \caption*{}
    \end{subfigure}
    \begin{subfigure}[b]{0.23\textwidth}
        \includegraphics[width=\textwidth]{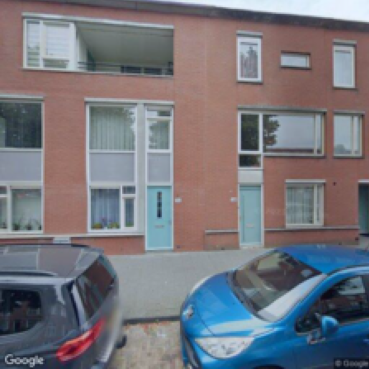}
        \caption*{}
    \end{subfigure}
    \begin{subfigure}[b]{0.23\textwidth}
        \includegraphics[width=\textwidth]{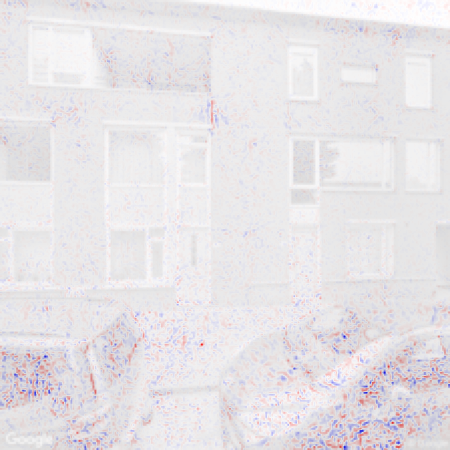}
        \caption*{}
    \end{subfigure}

        % Row for class "high and very high"
    \begin{subfigure}[b]{0.23\textwidth}
        \includegraphics[width=\textwidth]{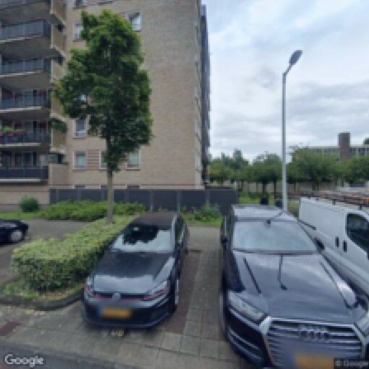}
        \caption*{high, very high risk}
    \end{subfigure}
    \begin{subfigure}[b]{0.23\textwidth}
        \includegraphics[width=\textwidth]{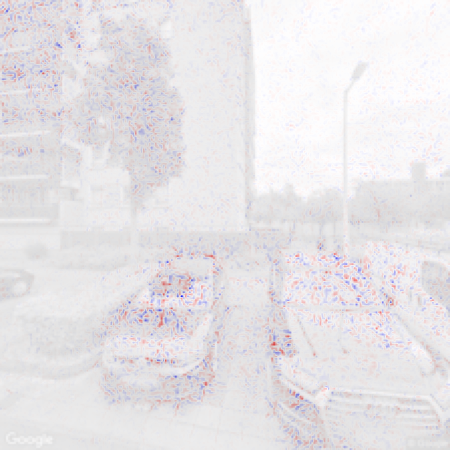}
        \caption*{}
    \end{subfigure}
    \begin{subfigure}[b]{0.23\textwidth}
        \includegraphics[width=\textwidth]{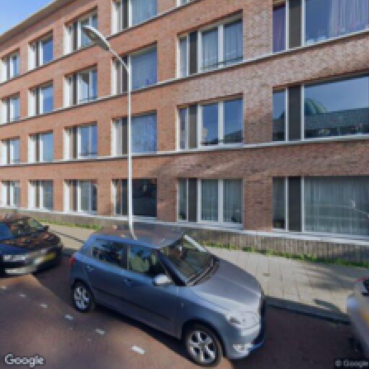}
        \caption*{}
    \end{subfigure}
    \begin{subfigure}[b]{0.23\textwidth}
        \includegraphics[width=\textwidth]{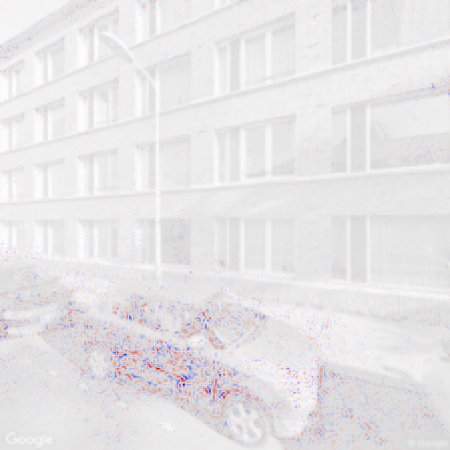}
        \caption*{}
    \end{subfigure}
    
    \caption{Visualization of SHAP values for the top-6 most certain predictions of ResNet50 per class. Part 2: moderate, high and very high risk. Images without overlay: Google Street View.}
    \label{fig:resnet_shap2}
\end{figure}

Figures \ref{fig:resnet_shap1} and \ref{fig:resnet_shap2} depict the regions of interest within the images that the model assigns the highest weights to, as revealed by the SHAP values per pixel. 

\subsubsection{DeiT.} 
\begin{figure}[htbp]
    \centering

    % Row for class "very low"
    \begin{subfigure}[b]{0.23\textwidth}
        \includegraphics[width=\textwidth]{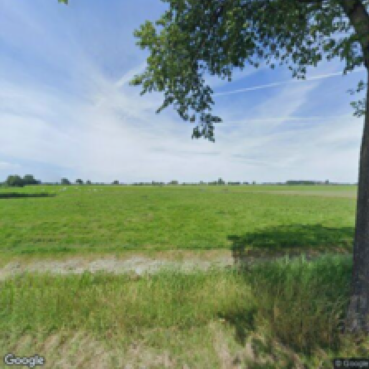}
        \caption*{}
    \end{subfigure}
    \begin{subfigure}[b]{0.23\textwidth}
        \includegraphics[width=\textwidth]{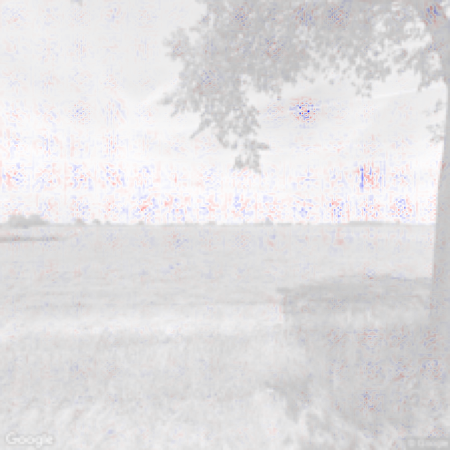}
        \caption*{}
    \end{subfigure}
    \begin{subfigure}[b]{0.23\textwidth}
        \includegraphics[width=\textwidth]{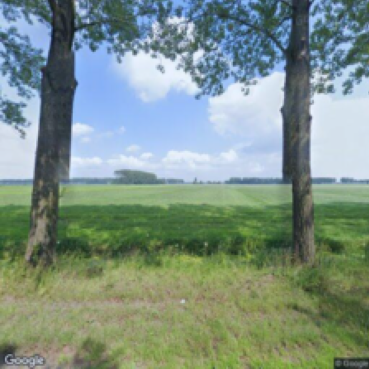}
        \caption*{}
    \end{subfigure}
    \begin{subfigure}[b]{0.23\textwidth}
        \includegraphics[width=\textwidth]{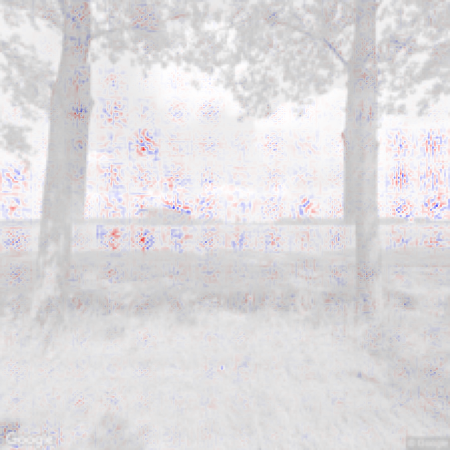}
        \caption*{}
    \end{subfigure}

        % Row for class "very low"
    \begin{subfigure}[b]{0.23\textwidth}
        \includegraphics[width=\textwidth]{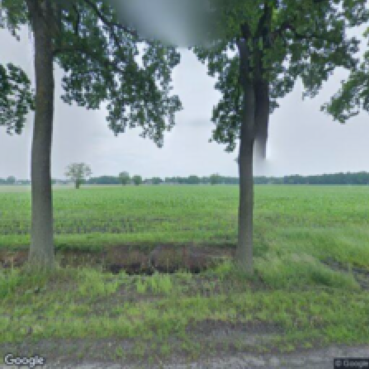}
        \caption*{}
    \end{subfigure}
    \begin{subfigure}[b]{0.23\textwidth}
        \includegraphics[width=\textwidth]{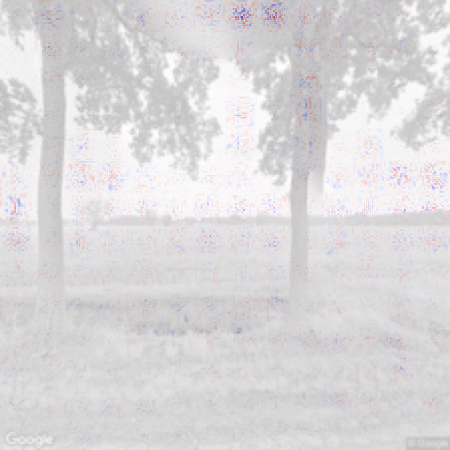}
        \caption*{}
    \end{subfigure}
    \begin{subfigure}[b]{0.23\textwidth}
        \includegraphics[width=\textwidth]{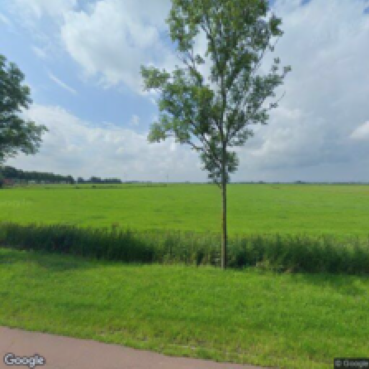}
        \caption*{}
    \end{subfigure}
    \begin{subfigure}[b]{0.23\textwidth}
        \includegraphics[width=\textwidth]{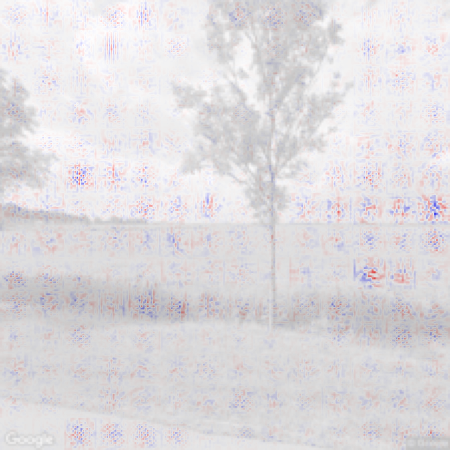}
        \caption*{}
    \end{subfigure}

        % Row for class "very low"
    \begin{subfigure}[b]{0.23\textwidth}
        \includegraphics[width=\textwidth]{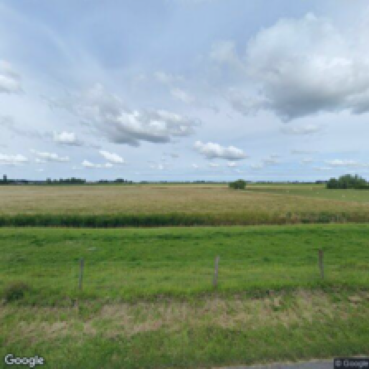}
        \caption*{very low risk}
    \end{subfigure}
    \begin{subfigure}[b]{0.23\textwidth}
        \includegraphics[width=\textwidth]{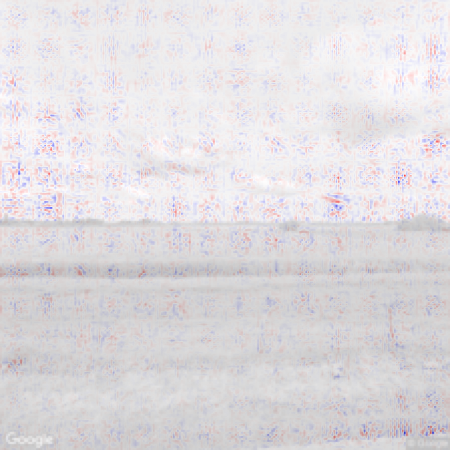}
        \caption*{}
    \end{subfigure}
    \begin{subfigure}[b]{0.23\textwidth}
        \includegraphics[width=\textwidth]{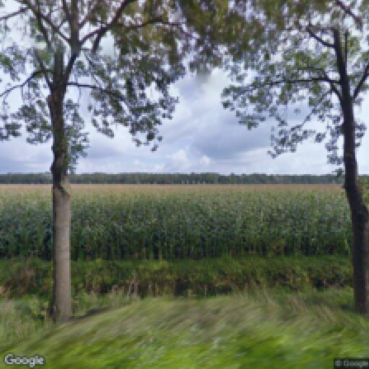}
        \caption*{}
    \end{subfigure}
    \begin{subfigure}[b]{0.23\textwidth}
        \includegraphics[width=\textwidth]{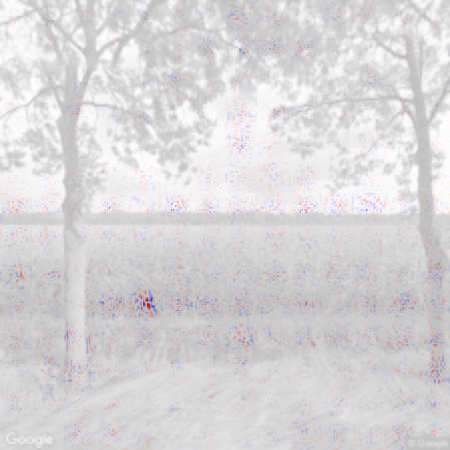}
        \caption*{}
    \end{subfigure}

       % Row for class "low"
    \begin{subfigure}[b]{0.23\textwidth}
        \includegraphics[width=\textwidth]{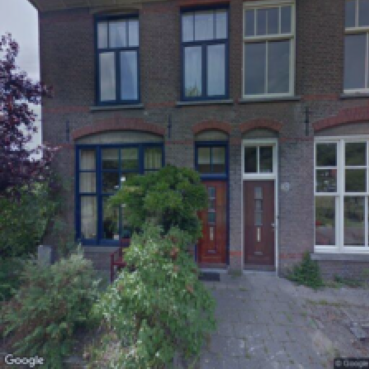}
        \caption*{}
    \end{subfigure}
    \begin{subfigure}[b]{0.23\textwidth}
        \includegraphics[width=\textwidth]{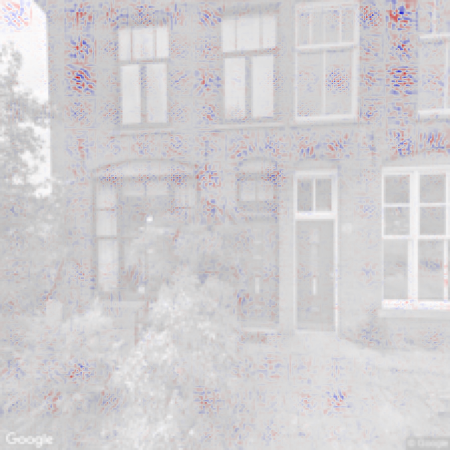}
        \caption*{}
    \end{subfigure}
    \begin{subfigure}[b]{0.23\textwidth}
        \includegraphics[width=\textwidth]{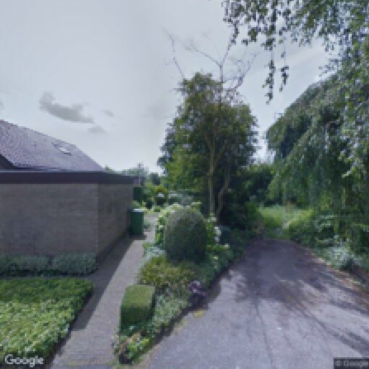}
        \caption*{}
    \end{subfigure}
    \begin{subfigure}[b]{0.23\textwidth}
        \includegraphics[width=\textwidth]{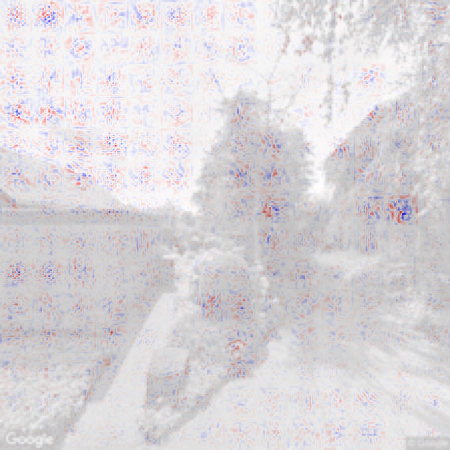}
        \caption*{}
    \end{subfigure}

        % Row for class "low"
    \begin{subfigure}[b]{0.23\textwidth}
        \includegraphics[width=\textwidth]{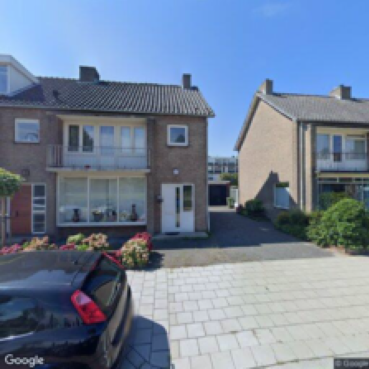}
        \caption*{}
    \end{subfigure}
    \begin{subfigure}[b]{0.23\textwidth}
        \includegraphics[width=\textwidth]{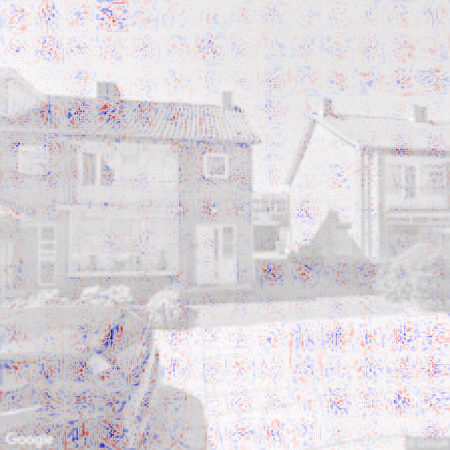}
        \caption*{}
    \end{subfigure}
    \begin{subfigure}[b]{0.23\textwidth}
        \includegraphics[width=\textwidth]{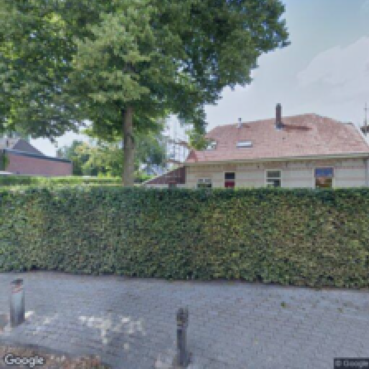}
        \caption*{}
    \end{subfigure}
    \begin{subfigure}[b]{0.23\textwidth}
        \includegraphics[width=\textwidth]{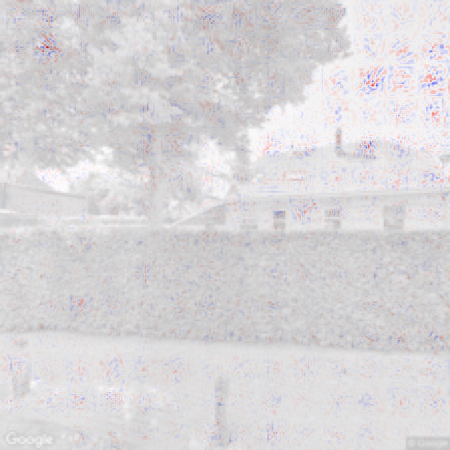}
        \caption*{}
    \end{subfigure}

        % Row for class "low"
    \begin{subfigure}[b]{0.23\textwidth}
        \includegraphics[width=\textwidth]{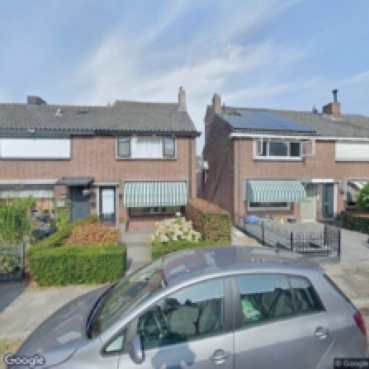}
        \caption*{low risk}
    \end{subfigure}
    \begin{subfigure}[b]{0.23\textwidth}
        \includegraphics[width=\textwidth]{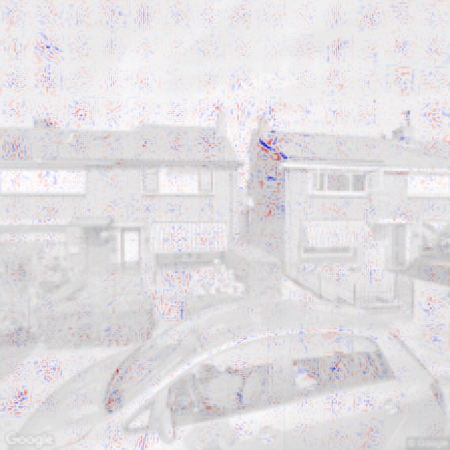}
        \caption*{}
    \end{subfigure}
    \begin{subfigure}[b]{0.23\textwidth}
        \includegraphics[width=\textwidth]{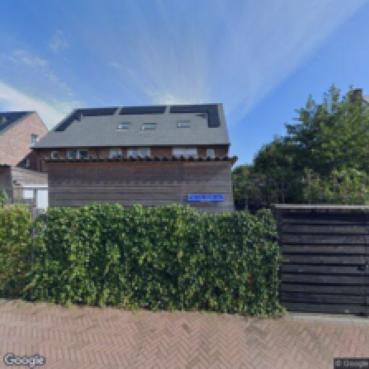}
        \caption*{}
    \end{subfigure}
    \begin{subfigure}[b]{0.23\textwidth}
        \includegraphics[width=\textwidth]{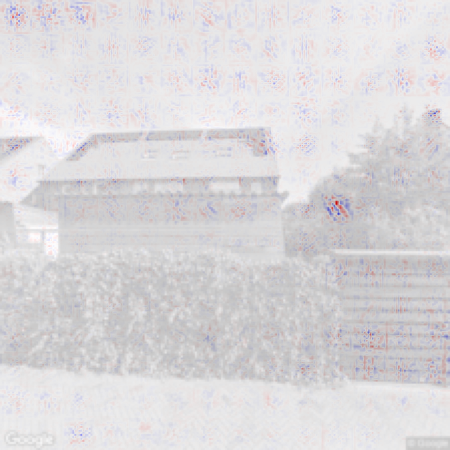}
        \caption*{}
    \end{subfigure}

    \caption{Visualization of SHAP values for the top-6 most certain predictions of DeiT Base per class. Part 1: very low and low risk. Images without overlay: Google Street View.}
    \label{fig:deit_shap1}
\end{figure}

\begin{figure}[htbp]
    \centering

           % Row for class "moderate"
    \begin{subfigure}[b]{0.23\textwidth}
        \includegraphics[width=\textwidth]{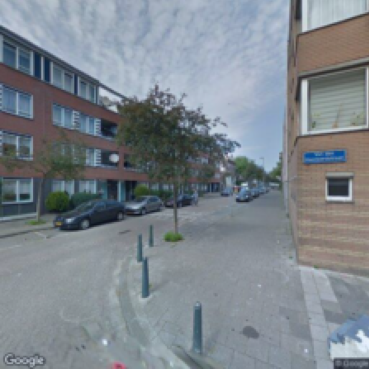}
        \caption*{}
    \end{subfigure}
    \begin{subfigure}[b]{0.23\textwidth}
        \includegraphics[width=\textwidth]{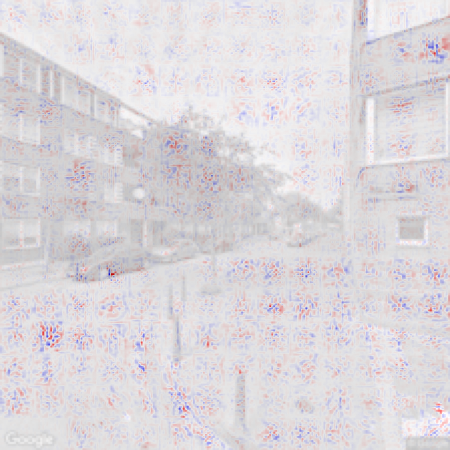}
        \caption*{}
    \end{subfigure}
    \begin{subfigure}[b]{0.23\textwidth}
        \includegraphics[width=\textwidth]{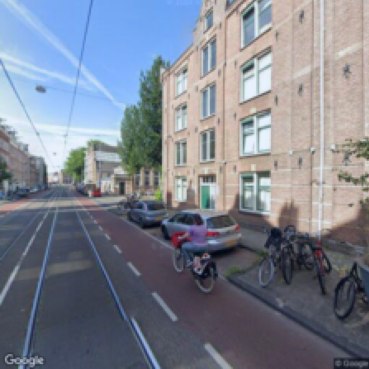}
        \caption*{}
    \end{subfigure}
    \begin{subfigure}[b]{0.23\textwidth}
        \includegraphics[width=\textwidth]{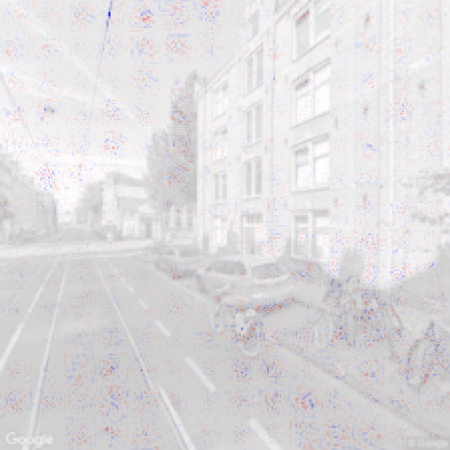}
        \caption*{}
    \end{subfigure}

        % Row for class "moderate"
    \begin{subfigure}[b]{0.23\textwidth}
        \includegraphics[width=\textwidth]{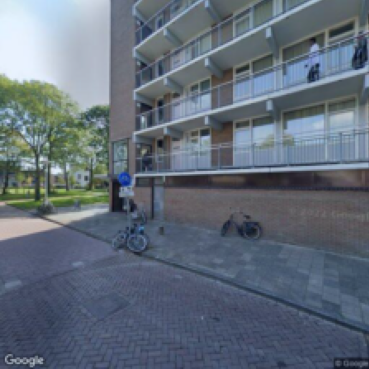}
        \caption*{}
    \end{subfigure}
    \begin{subfigure}[b]{0.23\textwidth}
        \includegraphics[width=\textwidth]{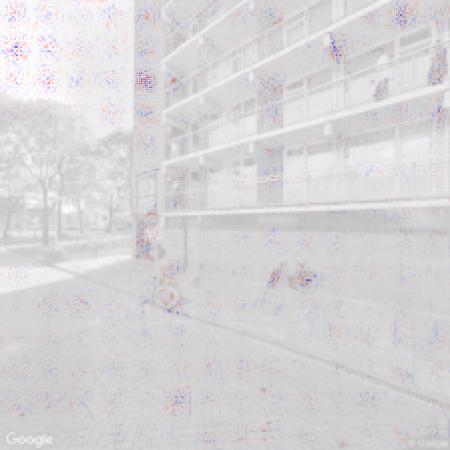}
        \caption*{}
    \end{subfigure}
    \begin{subfigure}[b]{0.23\textwidth}
        \includegraphics[width=\textwidth]{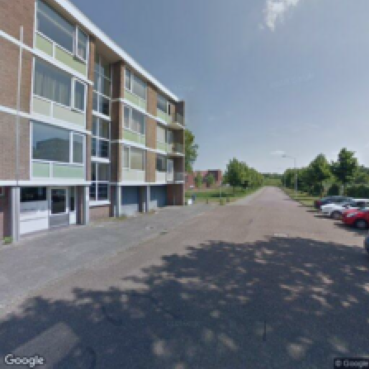}
        \caption*{}
    \end{subfigure}
    \begin{subfigure}[b]{0.23\textwidth}
        \includegraphics[width=\textwidth]{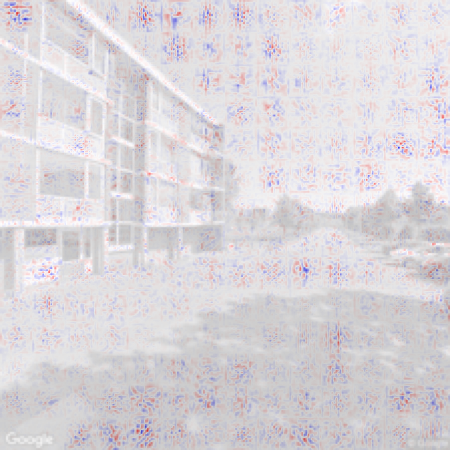}
        \caption*{}
    \end{subfigure}

     % Row for class "moderate"
    \begin{subfigure}[b]{0.23\textwidth}
        \includegraphics[width=\textwidth]{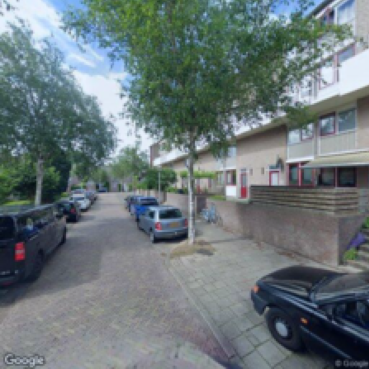}
        \caption*{moderate risk}
    \end{subfigure}
    \begin{subfigure}[b]{0.23\textwidth}
        \includegraphics[width=\textwidth]{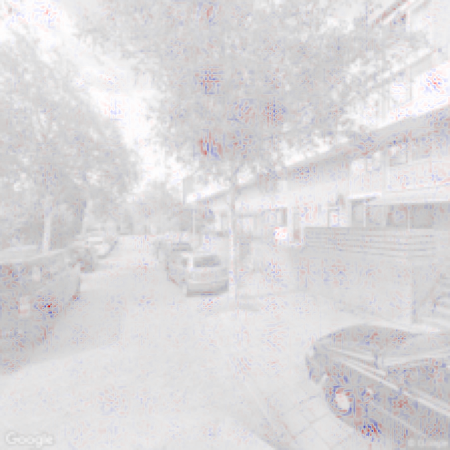}
        \caption*{}
    \end{subfigure}
    \begin{subfigure}[b]{0.23\textwidth}
        \includegraphics[width=\textwidth]{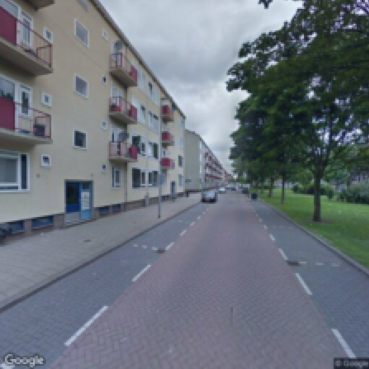}
        \caption*{}
    \end{subfigure}
    \begin{subfigure}[b]{0.23\textwidth}
        \includegraphics[width=\textwidth]{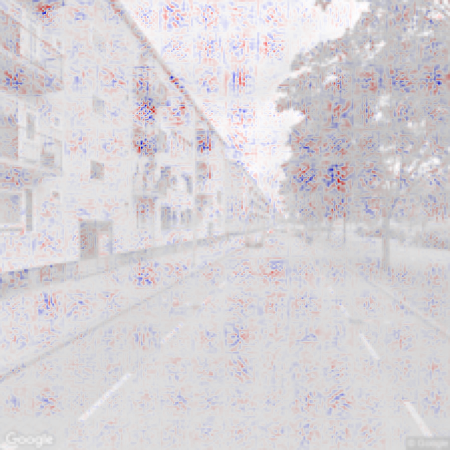}
        \caption*{}
    \end{subfigure}

           % Row for class "high and very high"
    \begin{subfigure}[b]{0.23\textwidth}
        \includegraphics[width=\textwidth]{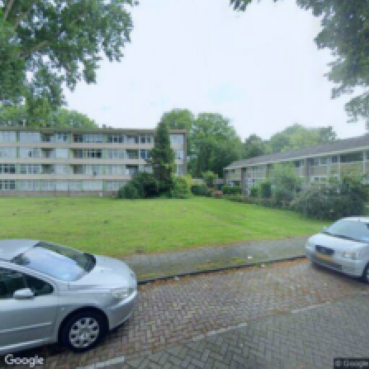}
        \caption*{}
    \end{subfigure}
    \begin{subfigure}[b]{0.23\textwidth}
        \includegraphics[width=\textwidth]{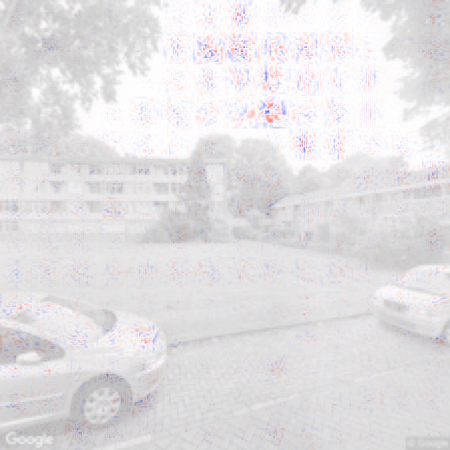}
        \caption*{}
    \end{subfigure}
    \begin{subfigure}[b]{0.23\textwidth}
        \includegraphics[width=\textwidth]{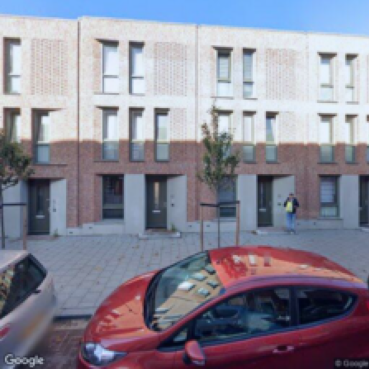}
        \caption*{}
    \end{subfigure}
    \begin{subfigure}[b]{0.23\textwidth}
        \includegraphics[width=\textwidth]{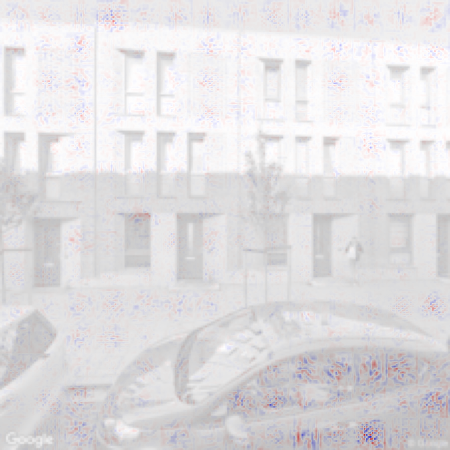}
        \caption*{}
    \end{subfigure}

        % Row for class "high and very high"
    \begin{subfigure}[b]{0.23\textwidth}
        \includegraphics[width=\textwidth]{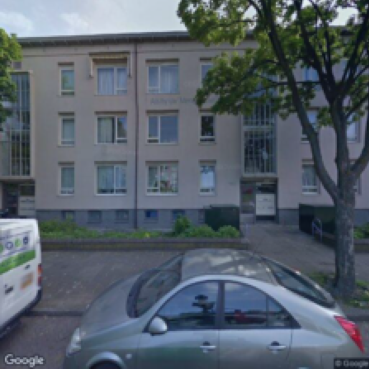}
        \caption*{}
    \end{subfigure}
    \begin{subfigure}[b]{0.23\textwidth}
        \includegraphics[width=\textwidth]{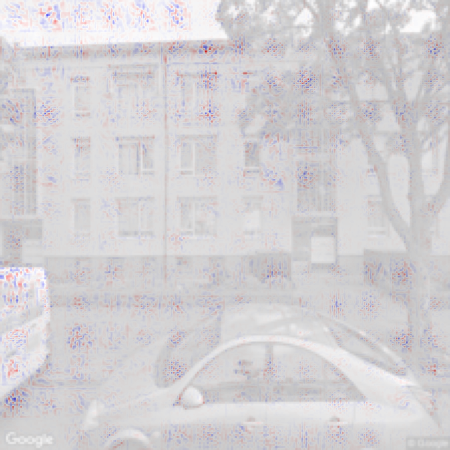}
        \caption*{}
    \end{subfigure}
    \begin{subfigure}[b]{0.23\textwidth}
        \includegraphics[width=\textwidth]{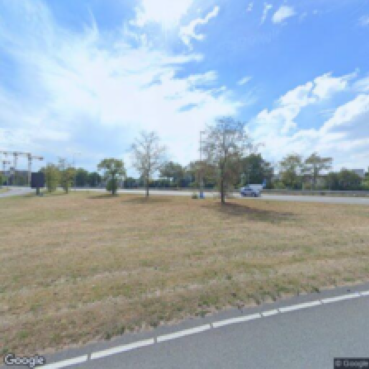}
        \caption*{}
    \end{subfigure}
    \begin{subfigure}[b]{0.23\textwidth}
        \includegraphics[width=\textwidth]{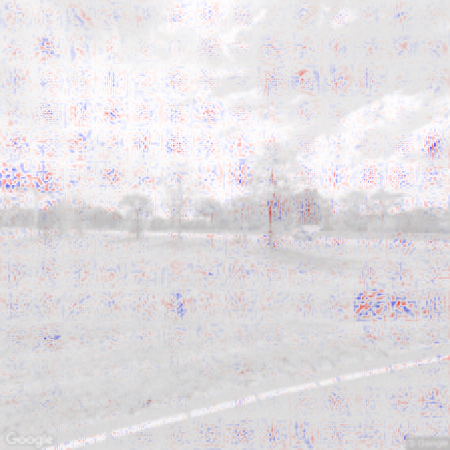}
        \caption*{}
    \end{subfigure}

        % Row for class "high and very high"
    \begin{subfigure}[b]{0.23\textwidth}
        \includegraphics[width=\textwidth]{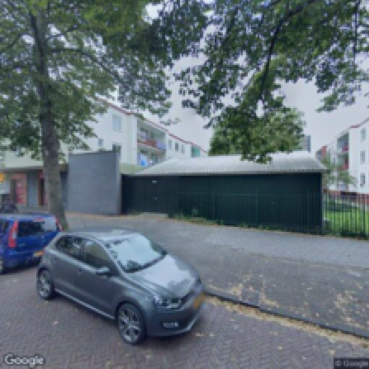}
        \caption*{high, very high risk}
    \end{subfigure}
    \begin{subfigure}[b]{0.23\textwidth}
        \includegraphics[width=\textwidth]{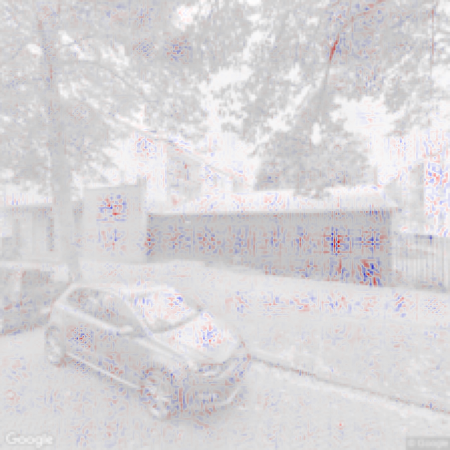}
        \caption*{}
    \end{subfigure}
    \begin{subfigure}[b]{0.23\textwidth}
        \includegraphics[width=\textwidth]{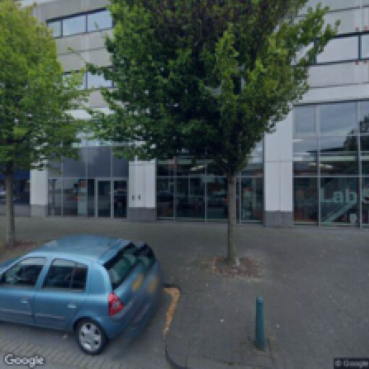}
        \caption*{}
    \end{subfigure}
    \begin{subfigure}[b]{0.23\textwidth}
        \includegraphics[width=\textwidth]{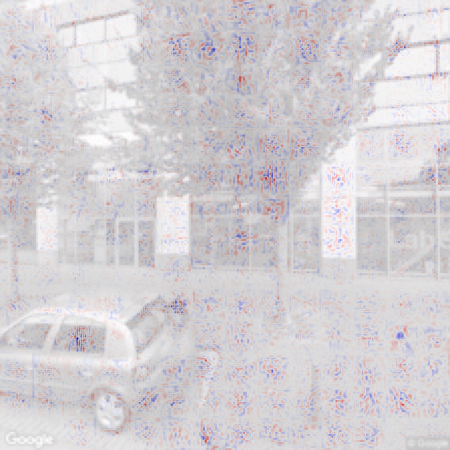}
        \caption*{}
    \end{subfigure}
    
    \caption{Visualization of SHAP values for the top-6 most certain predictions of DeiT Base per class. Part 2: moderate, high and very high risk. Images without overlay: Google Street View.}
    \label{fig:deit_shap2}
\end{figure}

Features highlighted by the visualization of the SHAP values (Figures \ref{fig:deit_shap1}, \ref{fig:deit_shap2}) are also ambiguent for this model. Among the ambiguously highlighted features the large areas of the sky are present again. For this model, however, the cars do not play an important role. As ResNet, DeiT seems to assign high importance to the vast sky areas and shadowy walls, based on the SHAP values. However, SHAP values are also extreme at the vast areas of pavement. SHAP visualizations also show that parked bicycles and motorbikes played an important role in classification for this model.

\begin{figure}[htbp]
    \centering

    \begin{tabularx}{\textwidth}{XXXX}
      "very low risk" & "low risk" & "low risk" \\
    \end{tabularx}

    \begin{subfigure}[b]{.3\textwidth}
        \centering
        \includegraphics[width=.9\linewidth]{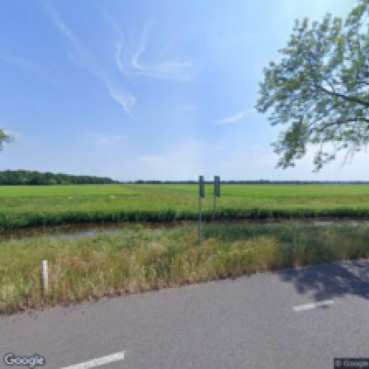}
        \caption*{}
        \label{fig:g1}
    \end{subfigure}%
    \begin{subfigure}[b]{.3\textwidth}
        \centering
        \includegraphics[width=.9\linewidth]{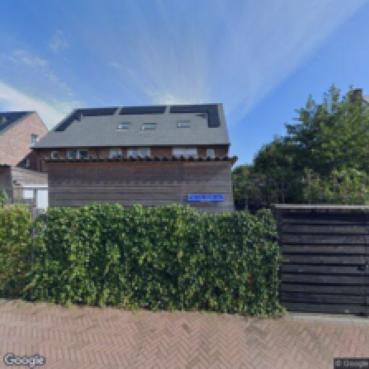}
        \caption*{}
        \label{fig:g2}
    \end{subfigure}
    \begin{subfigure}[b]{.3\textwidth}
        \centering
        \includegraphics[width=.9\linewidth]{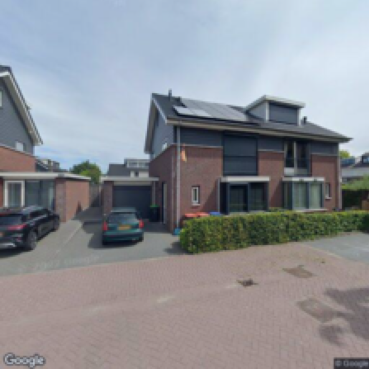}
        \caption*{}
        \label{fig:g3}
    \end{subfigure}%

    \begin{subfigure}[b]{.3\textwidth}
        \centering
        \includegraphics[width=.9\linewidth]{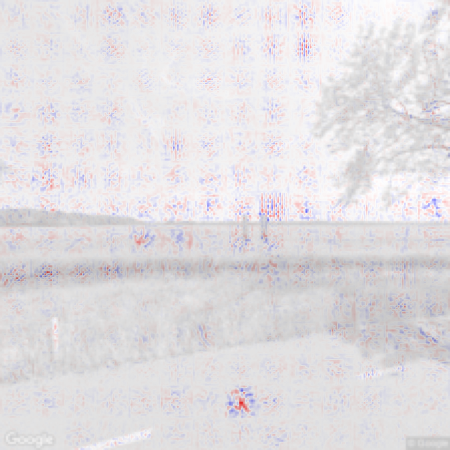}
        \caption*{(a)}
        \label{fig:g1}
    \end{subfigure}%
    \begin{subfigure}[b]{.3\textwidth}
        \centering
        \includegraphics[width=.9\linewidth]{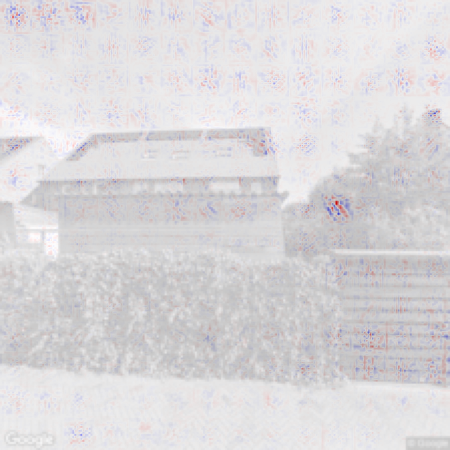}
        \caption*{(b)}
        \label{fig:g2}
    \end{subfigure}
    \begin{subfigure}[b]{.3\textwidth}
        \centering
        \includegraphics[width=.9\linewidth]{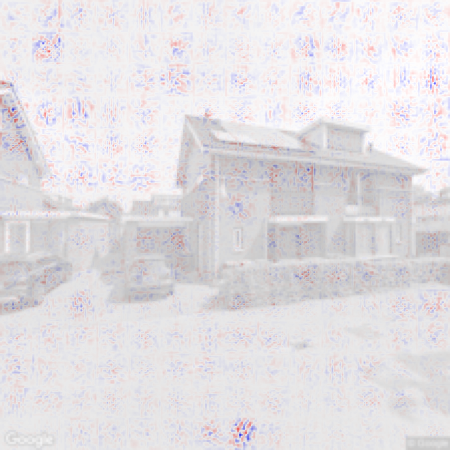}
        \caption*{(c)}
        \label{fig:g3}
    \end{subfigure}%

    \caption{Extreme SHAP values for DeiT Base predictions in unexpected places. Notice the blob of extreme values in the middle of the road surface in (a), on the seemingly monotonous canopy to the right in (b) and in the bottom of the seemingly monotonous pavement in (c). Images: Google Street View.}
    \label{fig:weird}
\end{figure}

The visualizations also exhibit sudden groups of extreme SHAP values in the places where no edges and no objects are present (Figure \ref{fig:weird}). 

\subsection{Attention maps}
\begin{figure}[htbp]
    \centering

    \begin{tabularx}{\textwidth}{XXXX}
    Original & Max & Min & Mean \\
    \end{tabularx}

    % Row for class "very low"
    \begin{subfigure}[b]{0.24\textwidth}
        \includegraphics[width=\textwidth]{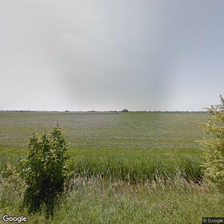}
        \caption*{very low}
    \end{subfigure}
    \begin{subfigure}[b]{0.24\textwidth}
        \includegraphics[width=\textwidth]{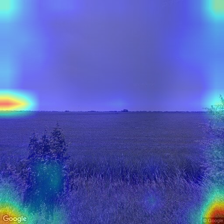}
        \caption*{}
    \end{subfigure}
    \begin{subfigure}[b]{0.24\textwidth}
        \includegraphics[width=\textwidth]{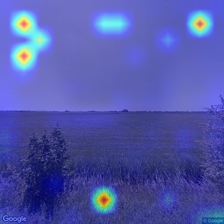}
        \caption*{}
    \end{subfigure}
    \begin{subfigure}[b]{0.24\textwidth}
        \includegraphics[width=\textwidth]{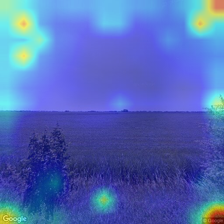}
        \caption*{}
    \end{subfigure}

    % Row for class "low"
    \begin{subfigure}[b]{0.24\textwidth}
        \includegraphics[width=\textwidth]{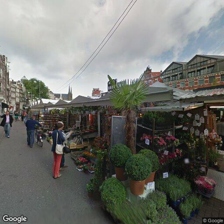}
        \caption*{low}
    \end{subfigure}
    \begin{subfigure}[b]{0.24\textwidth}
        \includegraphics[width=\textwidth]{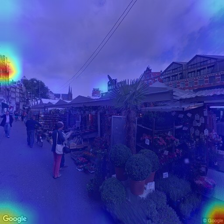}
        \caption*{}
    \end{subfigure}
    \begin{subfigure}[b]{0.24\textwidth}
        \includegraphics[width=\textwidth]{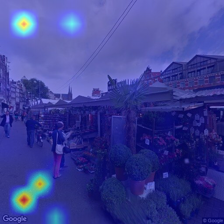}
        \caption*{}
    \end{subfigure}
    \begin{subfigure}[b]{0.24\textwidth}
        \includegraphics[width=\textwidth]{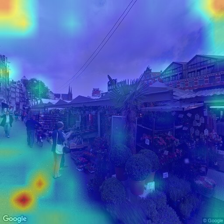}
        \caption*{}
    \end{subfigure}

    % Row for class "moderate"
    \begin{subfigure}[b]{0.24\textwidth}
        \includegraphics[width=\textwidth]{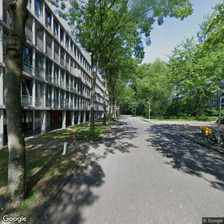}
        \caption*{moderate}
    \end{subfigure}
    \begin{subfigure}[b]{0.24\textwidth}
        \includegraphics[width=\textwidth]{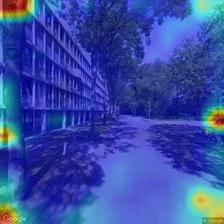}
        \caption*{}
    \end{subfigure}
    \begin{subfigure}[b]{0.24\textwidth}
        \includegraphics[width=\textwidth]{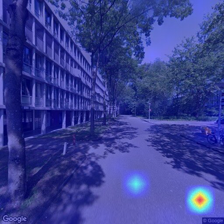}
        \caption*{}
    \end{subfigure}
    \begin{subfigure}[b]{0.24\textwidth}
        \includegraphics[width=\textwidth]{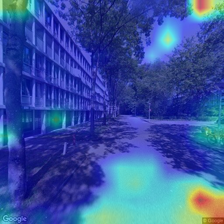}
        \caption*{}
    \end{subfigure}

    \begin{subfigure}[b]{0.24\textwidth}
        \includegraphics[width=\textwidth]{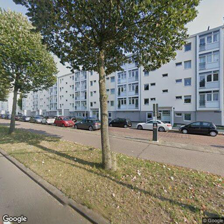}
        \caption*{high and very high}
    \end{subfigure}
    \begin{subfigure}[b]{0.24\textwidth}
        \includegraphics[width=\textwidth]{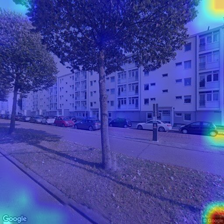}
        \caption*{}
    \end{subfigure}
    \begin{subfigure}[b]{0.24\textwidth}
        \includegraphics[width=\textwidth]{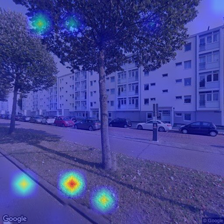}
        \caption*{}
    \end{subfigure}
    \begin{subfigure}[b]{0.24\textwidth}
        \includegraphics[width=\textwidth]{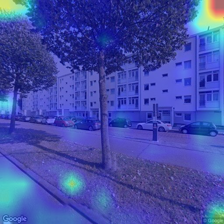}
        \caption*{}
    \end{subfigure}
    
    \caption{Attention maps for the most certain predictions of DeiT Base fully fine-tuned (with random crop). Each row represents a different class, with the columns showing the original image, and the top-20\% max, min, and mean attention weights respectively. Images without overlay: Google Street View.}
    \label{fig:fig10}
\end{figure}

\begin{figure}[htbp]
    \centering

    \begin{tabularx}{\textwidth}{XXXX}
    Original & Max & Min & Mean \\
    \end{tabularx}

    % Row for class "very low"
    \begin{subfigure}[b]{0.24\textwidth}
        \includegraphics[width=\textwidth]{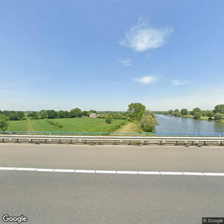}
        \caption*{very low}
    \end{subfigure}
    \begin{subfigure}[b]{0.24\textwidth}
        \includegraphics[width=\textwidth]{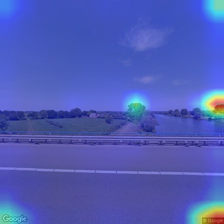}
        \caption*{}
    \end{subfigure}
    \begin{subfigure}[b]{0.24\textwidth}
        \includegraphics[width=\textwidth]{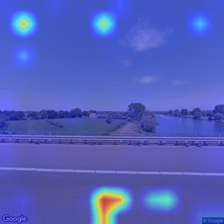}
        \caption*{}
    \end{subfigure}
    \begin{subfigure}[b]{0.24\textwidth}
        \includegraphics[width=\textwidth]{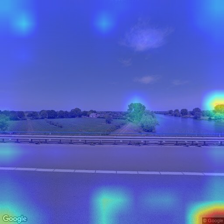}
        \caption*{}
    \end{subfigure}

    % Row for class "low"
    \begin{subfigure}[b]{0.24\textwidth}
        \includegraphics[width=\textwidth]{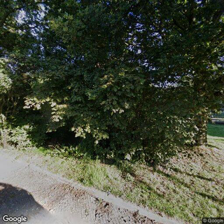}
        \caption*{low}
    \end{subfigure}
    \begin{subfigure}[b]{0.24\textwidth}
        \includegraphics[width=\textwidth]{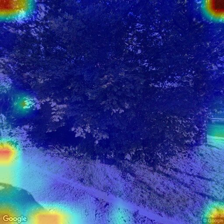}
        \caption*{}
    \end{subfigure}
    \begin{subfigure}[b]{0.24\textwidth}
        \includegraphics[width=\textwidth]{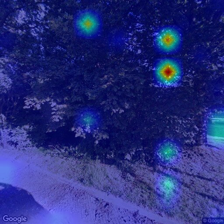}
        \caption*{}
    \end{subfigure}
    \begin{subfigure}[b]{0.24\textwidth}
        \includegraphics[width=\textwidth]{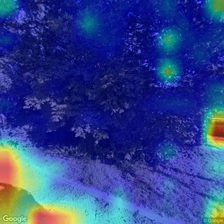}
        \caption*{}
    \end{subfigure}

    % Row for class "moderate"
    \begin{subfigure}[b]{0.24\textwidth}
        \includegraphics[width=\textwidth]{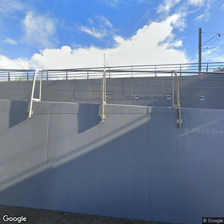}
        \caption*{moderate}
    \end{subfigure}
    \begin{subfigure}[b]{0.24\textwidth}
        \includegraphics[width=\textwidth]{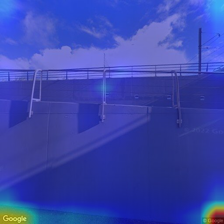}
        \caption*{}
    \end{subfigure}
    \begin{subfigure}[b]{0.24\textwidth}
        \includegraphics[width=\textwidth]{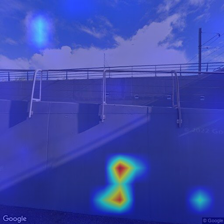}
        \caption*{}
    \end{subfigure}
    \begin{subfigure}[b]{0.24\textwidth}
        \includegraphics[width=\textwidth]{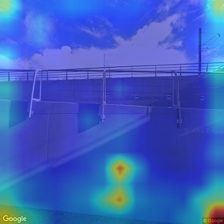}
        \caption*{}
    \end{subfigure}

    \begin{subfigure}[b]{0.24\textwidth}
        \includegraphics[width=\textwidth]{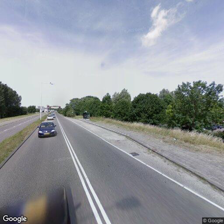}
        \caption*{high and very high}
    \end{subfigure}
    \begin{subfigure}[b]{0.24\textwidth}
        \includegraphics[width=\textwidth]{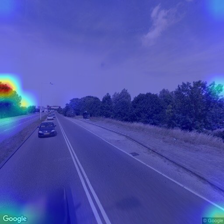}
        \caption*{}
    \end{subfigure}
    \begin{subfigure}[b]{0.24\textwidth}
        \includegraphics[width=\textwidth]{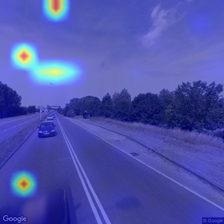}
        \caption*{}
    \end{subfigure}
    \begin{subfigure}[b]{0.24\textwidth}
        \includegraphics[width=\textwidth]{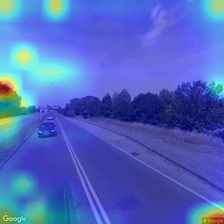}
        \caption*{}
    \end{subfigure}
    
    \caption{Attention maps for the least certain predictions of DeiT Base fully fine-tuned. Each row represents a different class, with the columns showing the original image, and the top-20\% max, min, and mean attention weights respectively. Images without overlay: Google Street View.}
    \label{fig:fig11}
\end{figure}

For a similar reason the visualization of the gradient rollout for DeiT Base seems to be uninformative to humans. The highlights are often in the middle of equally coloured field or sky area where no edges are present. Images produced for the top-10 most certain predictions were as uninformative as images produced for the bottom-10 least certain predictions (Figures \ref{fig:fig10}, \ref{fig:fig11}).

\end{document}